%% file: egpaper_for_review.tex
\documentclass[10pt,twocolumn,letterpaper]{article}

\usepackage{iccv}
\usepackage{times}
\usepackage{epsfig}
\usepackage{graphicx}
\usepackage{amsmath}
\usepackage{amssymb}
\usepackage{booktabs} 
\usepackage{soul}
\usepackage{enumitem}
\usepackage{tabularx}

\usepackage{color,colortbl}

\usepackage[dvipsnames]{xcolor}

\usepackage{pifont}
\newcommand{\cmark}{\ding{51}}%
\newcommand{\xmark}{\ding{55}}%
\definecolor{Gray}{gray}{0.9}
\definecolor{Gray2}{gray}{0.7}

\usepackage[pagebackref=true,breaklinks=true,letterpaper=true,colorlinks,bookmarks=false]{hyperref}

\iccvfinalcopy 

\def\iccvPaperID{11541} 

\begin{document}


\title{Exploring Open-Vocabulary Semantic Segmentation without Human Labels}

\author{Jun Chen\textsuperscript{\rm 1 \thanks{Work done while doing intern in Meta}}, Deyao Zhu\textsuperscript{\rm 1}, Guocheng Qian\textsuperscript{\rm 1}, Bernard Ghanem\textsuperscript{\rm 1},\\ Zhicheng Yan\textsuperscript{\rm 2}, Chenchen Zhu\textsuperscript{\rm 2}, Fanyi Xiao\textsuperscript{\rm 2}, Mohamed Elhoseiny\textsuperscript{\rm 1}, Sean Chang Culatana\textsuperscript{\rm 2}\\
\textsuperscript{\rm 1}{King Abdullah University of Science and Technology (KAUST)} \\
\textsuperscript{\rm 2}{Meta AI Research}\\
}


\maketitle

\begin{abstract}
Semantic segmentation is a crucial task in computer vision that involves segmenting images into semantically meaningful regions at the pixel level. 
However, existing approaches often rely on expensive human annotations as supervision for model training, 
limiting their scalability to large, unlabeled datasets. 
To address this challenge, we present ZeroSeg, 
a novel method that leverages the existing pretrained vision-language (VL) model (\eg CLIP~\cite{clip}) to train open-vocabulary zero-shot semantic segmentation models. 
Although acquired extensive knowledge of visual concepts, it is non-trivial to exploit knowledge from these VL models to the task of semantic segmentation, 
as they are usually trained at an image level. 
ZeroSeg overcomes this by distilling the visual concepts learned by VL models into a set of segment tokens,
each summarizing a localized region of the target image. 
We evaluate ZeroSeg on multiple popular segmentation benchmarks, 
including PASCAL VOC 2012, PASCAL Context, and COCO, in a zero-shot manner (\ie, no training or adaption on target segmentation datasets). 
Our approach achieves state-of-the-art performance when compared to other zero-shot segmentation methods under the same training data,
while also performing competitively compared to strongly supervised methods. 
Finally, we also demonstrated the effectiveness of ZeroSeg on open-vocabulary segmentation, through both human studies and qualitative visualizations.

\end{abstract}

\section{Introduction}
\input{Introduction}

\section{Related Works}
\input{Related_work}

\section{Method}
\input{Method}

\section{Results}
\input{Experiments}

\section{Discussion}
\input{Discussion}


{\small
\bibliographystyle{ieee_fullname}
\bibliography{egbib}
}

\clearpage
\appendix

\section{Supplementary material}
\subsection{HyperParameters}
The training hyperparamers are displayed in Table \ref{hyperparameter}.

\begin{table}[h!]
\begin{center}
\begin{tabular}{l|c}

config  & value \\
\toprule
optimizer & AdamW \\
base learning rate &  1e-4 \\
weight decay  &  0.05 \\
optimizer momentum & $\beta_1$, $\beta_2$=0.9,0.95 \\
batch size &  4096 \\
total epochs & 80 \\
warmup epochs & 20 \\
masked decoder layer & 8 \\
first-stage grouping layer & 2 \\
second-stage grouping layers & 2\\
first-stage group tokens  & 32 \\
second-stage group tokens  & 8 \\
\end{tabular}
\end{center}
\caption{
\textbf{Hyperparamer setting}.
} 
\label{hyperparameter}
\end{table}

\subsection{Training compute}
\noindent \textbf{Comparison of Training Efficiency with GroupViT:} The Table \ref{compute} demonstrates that GroupViT requires about 768 V100 training hours to complete 30 epochs, whereas ZeroSeg only requires around 84
training hours for 80 epochs. This means that  ZeroSeg achieves comparable results with only = an order of magnitude less computational resources compared to GroupViT (1/9), demonstrating its training efficiency

\begin{table}[h!]
\setlength{\tabcolsep}{3pt}
\begin{center}
\begin{tabular}{lcc}
  & GroupViT & ZeroSeg \\
\toprule
GPU hours & $\sim$768 & $\sim$84 \\
epochs & 30 & 80  \\ 
\end{tabular}
\end{center}
\caption{V100 GPU compute comparison}
\label{compute}
\end{table}

\subsection{Performance scaling with larger amount of training dataset.}
\begin{table}[h!]
\setlength{\tabcolsep}{3pt}
\begin{center}
\begin{tabular}{lcccc}
Model & Datasets & VOC & Context & COCO  \\
\toprule
GroupViT & CC12M  & 41.1 & - & - \\
ZeroSeg & IN-1K & 40.8 & 20.4 & 20.2  \\
ZeroSeg & CC12M + IN-1K & \textbf{42.9} & \textbf{21.8} & \textbf{22.1} \\
\end{tabular}
\end{center}
\caption{Training with larger-scaled datasets.} 

\label{performance_scaling}
\end{table}

To assess the scalability of our model, we conducted experiments with larger datasets. 
Due to the inaccessibility of the YFCC100M dataset, we had to seek alternatives by training ZeroSeg using CC12M+IN-1K datasets instead. 
The results, presented in Table \ref{performance_scaling}, demonstrated that scaling up the training data leads to improved performance for ZeroSeg. 
Notably, ZeroSeg (CC12M+IN-1K) outperforms GroupViT (CC12M) by 1.8 mIoU on VOC dataset, providing further evidence of the scalability of our model.

\subsection{More visualization examples}
\begin{figure*}[t!]
\begin{minipage}{0.49\textwidth}

\centering
 \includegraphics[width=1.0\linewidth]
    {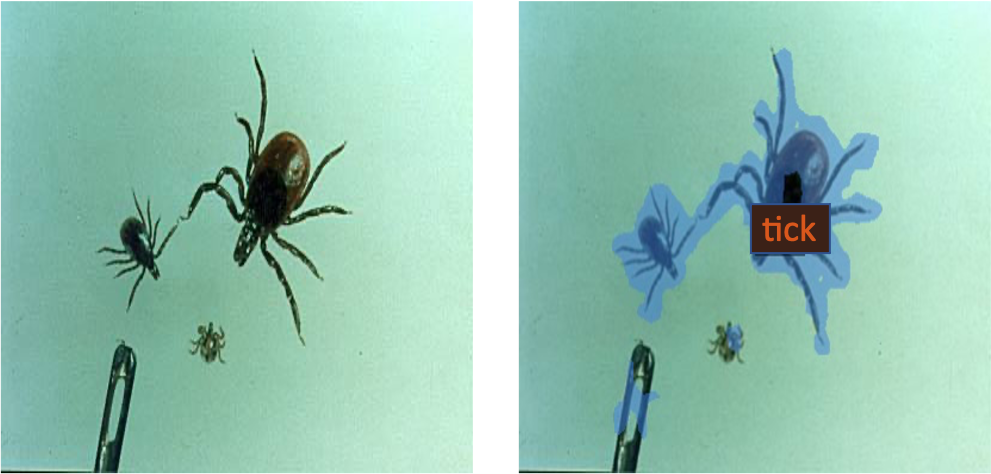}   
\end{minipage} \hfill
\begin{minipage}{0.49\textwidth}
\centering
 \includegraphics[width=1.0\linewidth]
    {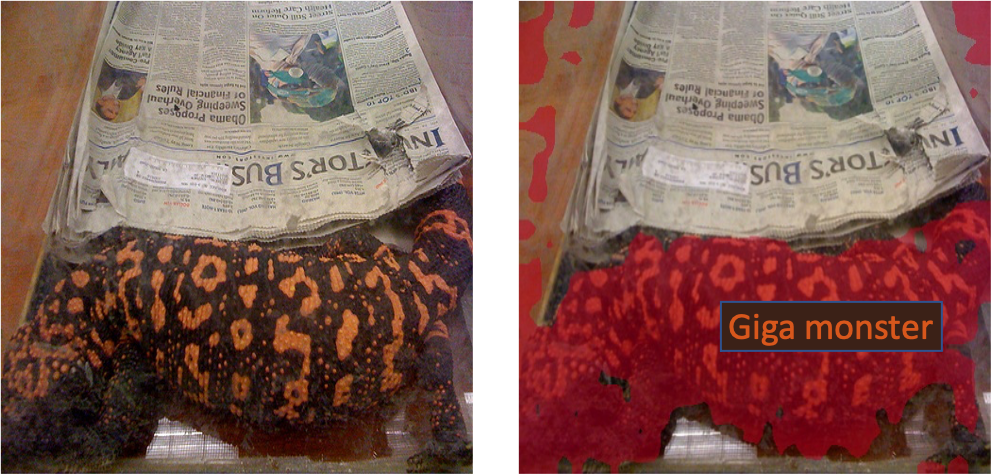}   
\end{minipage} \hfill

\begin{minipage}{0.495\textwidth}

\centering
 \includegraphics[width=1.0\linewidth]
    {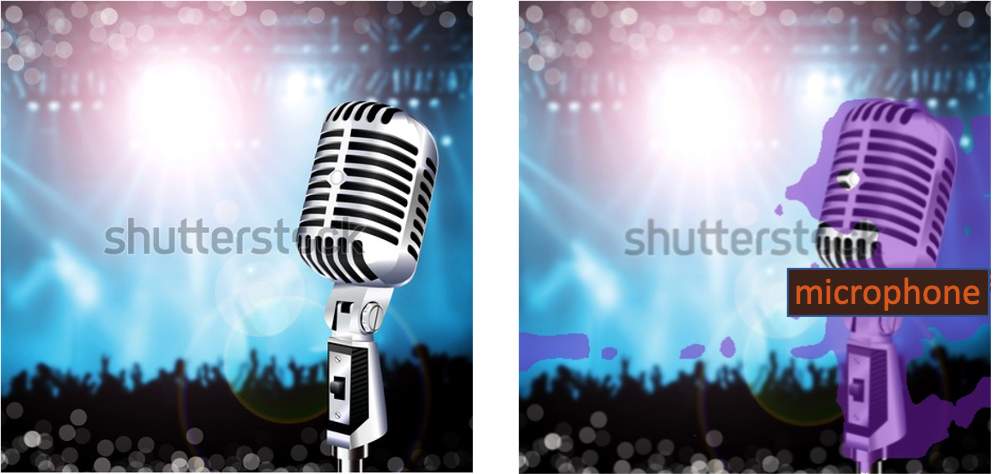}   
\end{minipage} \hfill
\begin{minipage}{0.49\textwidth}
\centering
 \includegraphics[width=1.0\linewidth]
    {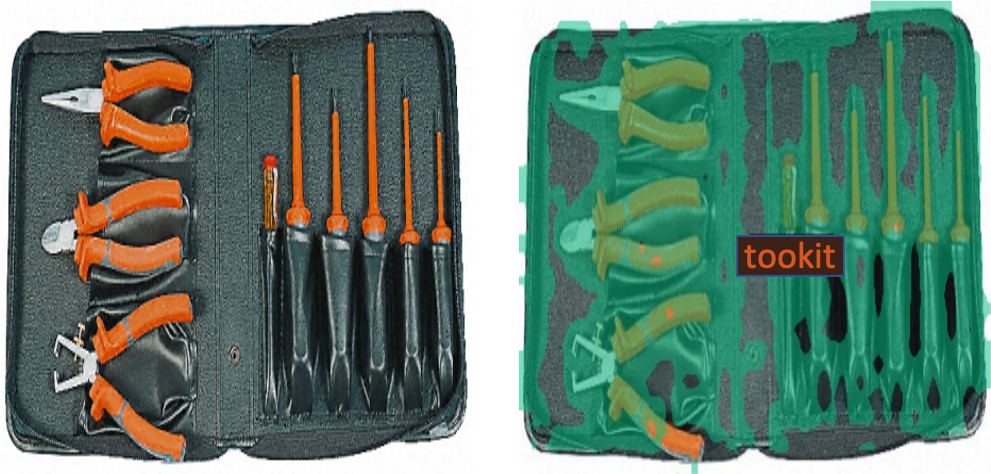}   
\end{minipage} \hfill

\begin{minipage}{0.49\textwidth}

\centering
 \includegraphics[width=1.0\linewidth]
    {figures/supp\_cc1.png}   
\end{minipage} \hfill
\begin{minipage}{0.49\textwidth}
\centering
 \includegraphics[width=1.0\linewidth]
    {figures/supp\_cc2.png}   
\end{minipage} \hfill

\begin{minipage}{0.49\textwidth}
\centering
 \includegraphics[width=1.0\linewidth]
    {figures/supp\_cc3.png} 
    \title{ Input  \hspace{90pt} ZeroSeg}
\end{minipage} \hfill
\begin{minipage}{0.49\textwidth}
\centering
 \includegraphics[width=1.0\linewidth]
    {figures/supp\_cc4.png}   
    \title{Input  \hspace{90pt} ZeroSeg}
\end{minipage} \hfill
\vspace{1em}
\caption{
\textbf{More sampled example from ImageNet and Conceptual Caption val set
}}
\label{fig1}
\end{figure*}

\begin{figure*}[t!]
\begin{minipage}{0.49\textwidth}
\centering
 \includegraphics[width=1.0\linewidth]
    {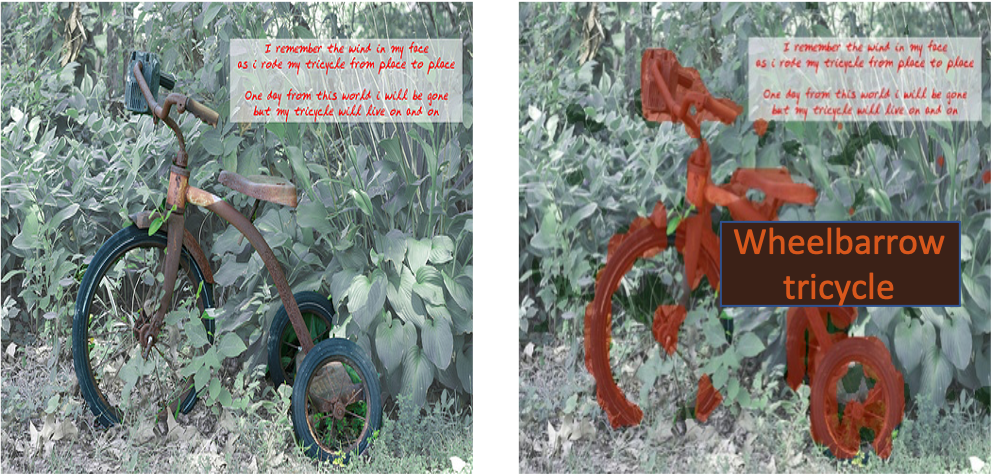}   
\end{minipage} \hfill
\begin{minipage}{0.49\textwidth}
\centering
 \includegraphics[width=1.0\linewidth]
    {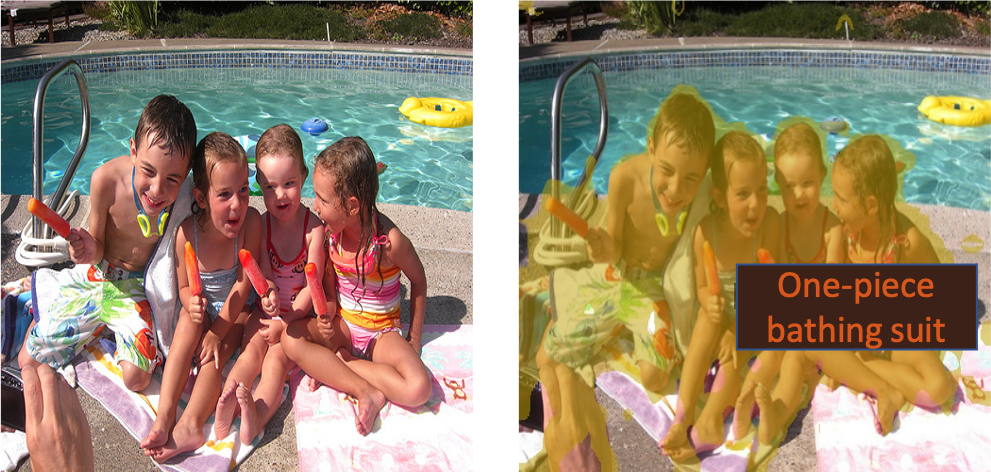}   
\end{minipage} \hfill
\begin{minipage}{0.49\textwidth}
\centering
 \includegraphics[width=1.0\linewidth]
    {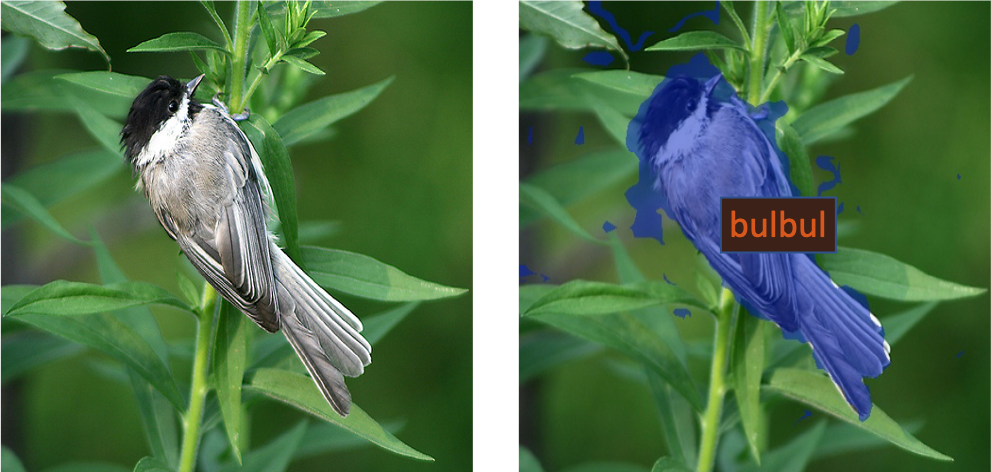} 
\end{minipage} \hfill
\begin{minipage}{0.49\textwidth}
\centering
 \includegraphics[width=1.0\linewidth]
    {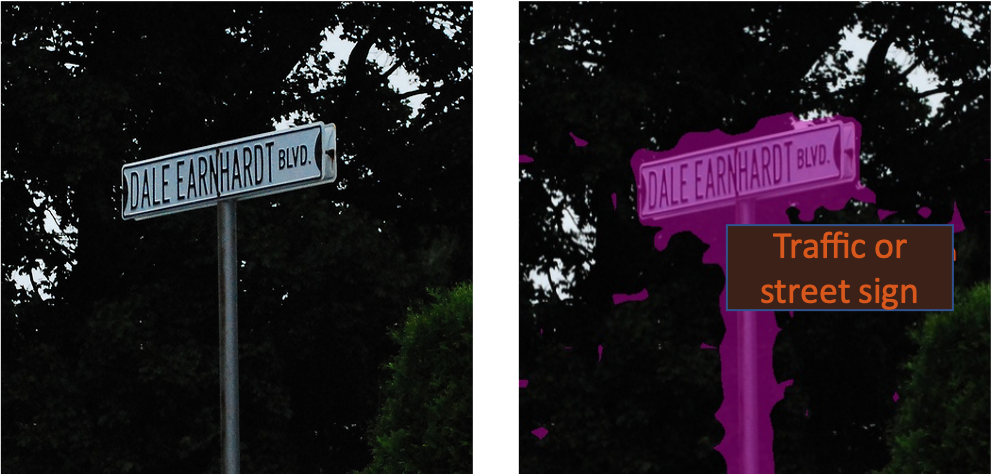}   
\end{minipage} \hfill

\begin{minipage}{0.495\textwidth}
\centering
 \includegraphics[width=1.0\linewidth]
    {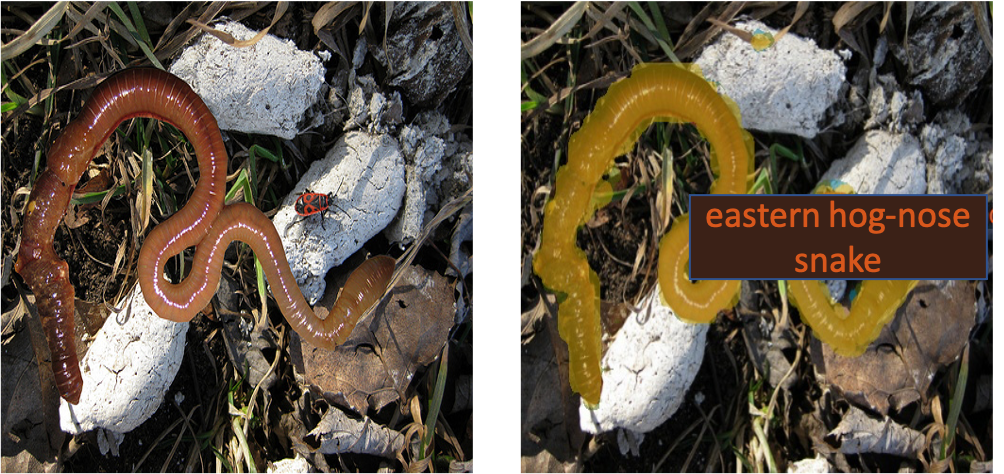}   
\end{minipage} \hfill
\begin{minipage}{0.495\textwidth}
\centering
 \includegraphics[width=1.0\linewidth]
    {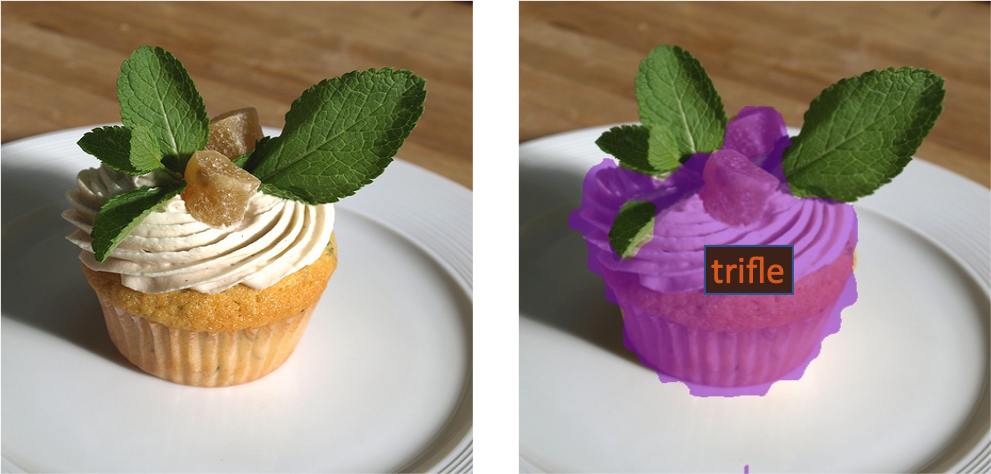}   
\end{minipage} \hfill
\begin{minipage}{0.49\textwidth}
\centering
 \includegraphics[width=1.0\linewidth]
    {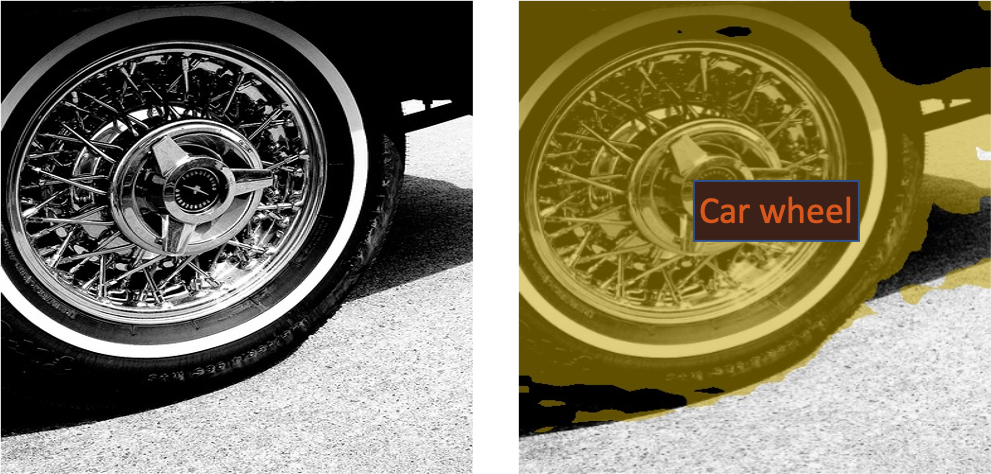} 
    \title{ Input \hspace{90pt} ZeroSeg}
\end{minipage} \hfill
\begin{minipage}{0.49\textwidth}
\centering
 \includegraphics[width=1.0\linewidth]
    {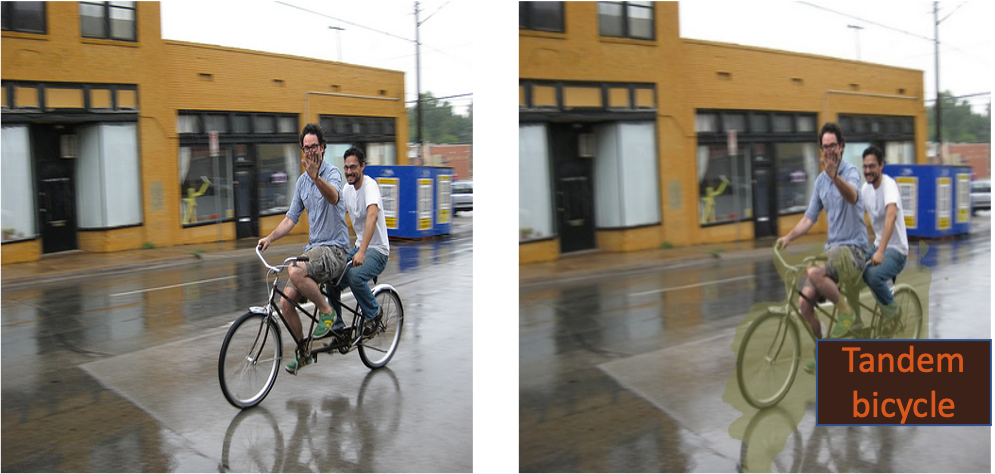}   
    \title{Input    \hspace{90pt} ZeroSeg}
\end{minipage} \hfill
\vspace{1em}
\caption{
\textbf{More sampled example from ImageNet and Conceptual Caption val set
}}
\label{fig2}
\end{figure*}

We sample more examples from ImageNet 1k \cite{imagenet} and Conceptual Caption \cite{conceptual} val dataset and demonstrate them in the Fig. \ref{fig1} and Fig. \ref{fig2}
\clearpage

\end{document}

%% file: Introduction.tex

Semantic segmentation involves dividing an image into distinct regions and assigning each area a corresponding label, 
and the open-vocabulary setting targets performing segmentation with an unrestricted vocabulary. 
This process typically necessitates human-generated annotations, 
such as per-pixel label supervision \cite{zhao2017pyramid,fu2019dual,huang2019ccnet,strudel2021segmenter,wang2018non,yuan2021ocnet,zheng2021rethinking,cheng2022masked}, 
or image-level supervision, e.g. human natural language \cite{openseg,dong2022maskclip,groupvit}. 
However, it can be time-consuming and expensive to obtain these annotations, 
and thus the resulting model can not be trained on large amounts of data. 
Recently, new developments in the field of vision and language learning \cite{clip,li2023blip,flamingo,yuan2021florence,chen2022visualgpt,zhu2023minigpt} have emerged. 
Although some of these approaches have demonstrated impressive open-vocabulary image/object classification capabilities, 
their performance for open-vocabulary semantic segmentation has been less promising. 
Nonetheless, they provide a potential alternative solution to overcome the limitations of traditional supervised methods.

\begin{figure}[t!]
\centering
 \includegraphics[width=1.0\linewidth,height =0.8\linewidth]
    {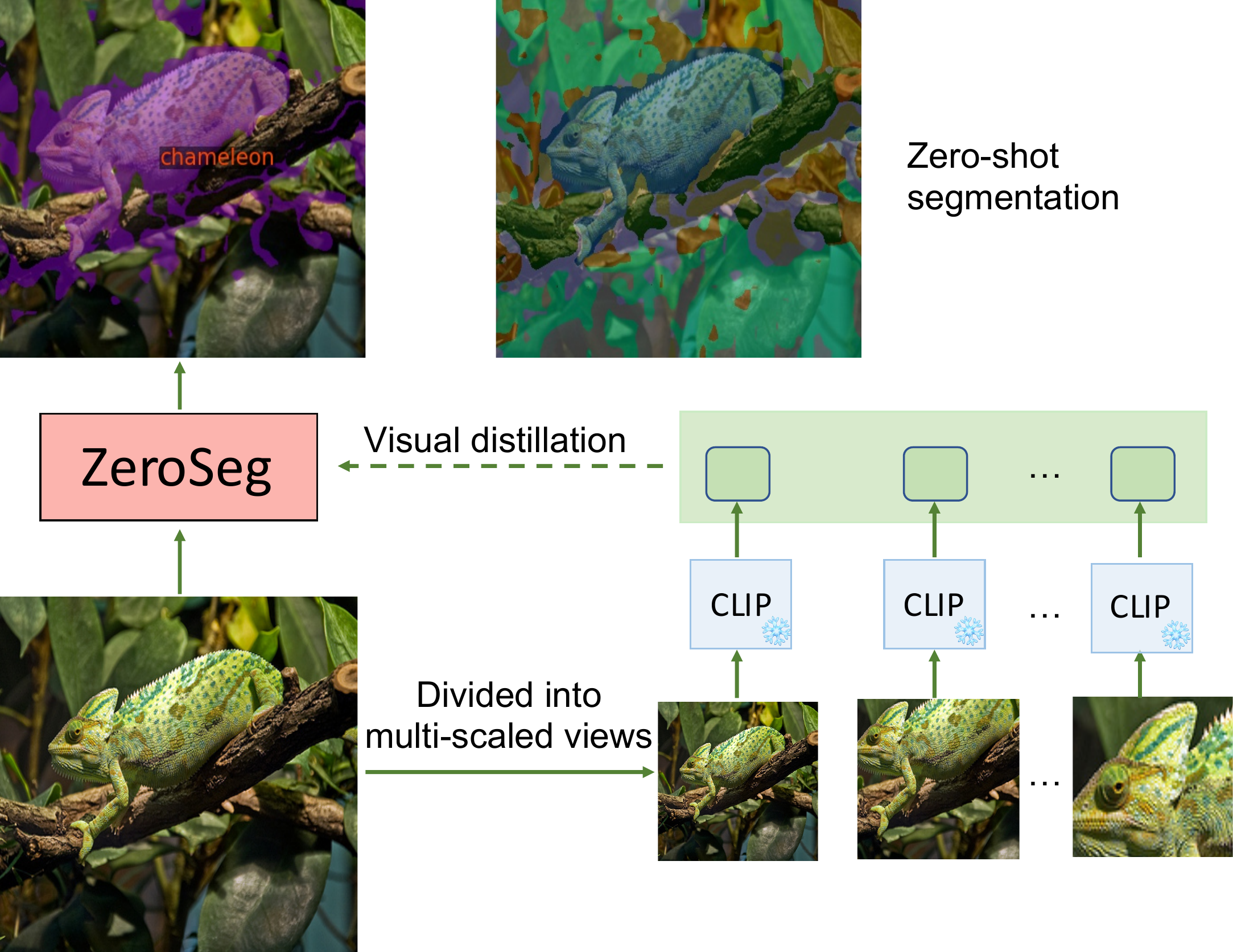}   
\caption{\textbf{ZeroSeg overview.} 
ZeroSeg is a zero-shot open-vocabulary method for semantic segmentation. 
The approach begins by dividing the input image into a set of multi-scale views. 
Each view is then individually processed by a pretrained CLIP visual encoder model to extract visual concepts. 
These visual concepts are then distilled into our ZeroSeg model via the proposed segment matching loss. 
After training, our ZeroSeg model can be directly transferred to downstream semantic segmentation tasks in a zero-shot manner (\ie, no training or adaption on target datasets). 
The entire training process does not require any human labels. 
}
\label{overview}
\end{figure}

To improve the scalability of semantic segmentation for a large or open vocabulary, researchers have explored models that can learn directly from tens of millions of text samples~\cite{openseg,groupvit,dong2022maskclip}. 
However, these vision-language (VL) models are prohibitively expensive to train and thus it is best to be able to exploit pretrained VL model weights (\eg, CLIP) for downstream segmentation tasks.  
However, to directly adapt CLIP for per-pixel semantic segmentation is not trivial, since CLIP has only been trained using coarse-grained image-level supervision, even though it has learned extensive visual concepts. 

Initial attempts have been made to also leverage pretrained vision-language models for open-vocabulary semantic segmentation, such as those discussed in \cite{xu2022simple,segclip}. 
However, these previous attempts primarily treated CLIP as a zero-shot segment-level classifier or as a visual backbone for the improved initialization. 
They usually still need to require expensive per-pixel level labels or extensive image-text pairs for the training.
In contrast, our proposed method treats CLIP as a teacher model and distills its knowledge into our newly designed segmentation model, named ZeroSeg, to facilitate semantic segmentation. 
This process enables the direct transfer of various learned visual concepts into ZeroSeg without the need for any human labels, thereby naturally extending CLIP for open-vocabulary semantic segmentation.

One of the main challenges in using a large pretrained vision-language model for per-pixel level supervision is how to effectively group and categorize semantically consistent pixels. To tackle this problem, we have incorporated a segments-grouping approach \cite{groupvit} into our ZeroSeg model. This approach automates the grouping of pixels into more significant, arbitrary-shaped segments. With these segments, it then becomes much easier to distill semantic information from the CLIP visual encoder to these localized image regions. As illustrated in Fig. \ref{overview}, ZeroSeg divides the input image into multiple scaled regions and extracts their semantic features via the CLIP visual encoder. Each of those regional features will be distilled into a set of learnable segment tokens both locally and globally. The visual segments will finally emerge to match the consistency with the different scales of semantic information from CLIP. Additionally, to improve the efficiency of training, our model also incorporates a masked autoencoder \cite{mae}.


To assess the efficacy of our proposed model, we trained ZeroSeg using only the ImageNet 1k dataset \cite{imagenet}, without any human label supervision. Our findings reveal that our model is comparable in performance to those that were trained with human-label supervision. Specifically, we achieved a mean intersection over union (mIoU) of 40.8 on PASCAL VOC 2012 \cite{voc}, a mIoU of 20.6 on PASCAL Context \cite{context}, and a mIoU of 20.4 on the COCO dataset \cite{coco} in a zero-shot manner. These results are comparable to models such as GroupViT \cite{groupvit} and MaskCLIP \cite{dong2022maskclip}, which were pretrained on 26M and 20M image-text pairs, respectively, indicating the efficiency and effectiveness of our approach. Additionally, our model has performed well in a larger-vocabulary (1000 classes) semantic segmentation task. Our work is the first to enable open-vocabulary semantic segmentation by only distilling knowledge from the pretrained vision-language without any human labels.

\noindent \textbf{Contributions.} We make the following contributions:
 \begin{itemize}

     \item Our research introduces ZeroSeg, a model that enables efficient open-vocabulary semantic segmentation without relying on human annotations. By distilling knowledge from a pretrained vision-language model, ZeroSeg bypasses the need for training on a large dataset of image-text pairs.
    
    \item The success of ZeroSeg is attributed to its carefully-designed architecture, which includes segment matching loss and multi-scaled feature distillation loss. These components are crucial for achieving accurate segmentation without human labels.
    
    \item Despite being pretrained on only ImageNet-1k, which has almost 20 times fewer samples than the other baseline models trained on text supervision, ZeroSeg achieves comparable results. As a result, our model provides a significant increase in training efficiency without sacrificing performance.

 \end{itemize}

%% file: Related_work.tex

\noindent \textbf{Supervised semantic segmentation.} Fully supervised semantic segmentation methods rely on per-pixel level supervision and have achieved significant success. Many such methods have been proposed, including \cite{chen2017rethinking,long2015fully,zhao2017pyramid,fu2019dual,huang2019ccnet,strudel2021segmenter,wang2018non,yuan2021ocnet,zheng2021rethinking,cheng2022masked}. They have achieved strong performance for in-domain semantic segmentation. However, these methods often struggle to generalize to new visual concepts that were not present in the training dataset. This limitation can be attributed to the fact that fully supervised methods require pixel-level annotations for all object classes of interest, making them impractical for scenarios where new object classes are encountered at test time.

\noindent \textbf{Semantic segmentation with less supervision.} Obtaining dense per-pixel labels is often costly and time-consuming, leading to a trend of research on learning to segment with less supervision. Some works leverage image-level labels, such as classification labels \cite{Wang_2020_CVPR,pinheiro2015image,Xu_2021_ICCV}, captions \cite{openseg,dong2022maskclip,groupvit}, or pseudo-masks \cite{li2021pseudo}. Few-shot methods \cite{lu2021simpler,dong2018few,liu2020part,nguyen2019feature,tian2020prior,yang2021mining} have also been proposed to perform segmentation with fewer pixel-wise labels. In addition, zero-shot semantic segmentation
approaches~\cite{bucher2019zero,xian2019semantic,hu2020uncertainty,baek2021exploiting,li2020consistent} have been developed to segment unseen visual concepts by aligning with language embeddings, but they still require per-pixel label supervision on seen categories at the beginning. Our approach differs from previous methods in that we rely solely on a CLIP vision encoder as the teacher without any per-pixel labels or language signals as supervision, allowing our strategy to train on any images. This enables more flexible and efficient semantic segmentation learning.

\noindent \textbf{Open-vocabulary segmentation.} Open-vocabulary segmentation aims to segment images beyond a closed-set vocabulary. Early attempts at open-vocabulary segmentation involved linking pixels to word concepts from WordNet \cite{zhao2017open}. However, recent developments in CLIP-based methods have significantly improved the ability to perform open-vocabulary segmentation. For example, Xu \textit{et al.} \cite{xu2022simple} propose using CLIP to classify mask segments generated by a pretrained mask generator \cite{cheng2021per}. Li \textit{et al.} encode pixel embeddings from a pretrained visual encoder and classify each embedding with the CLIP text encoder \cite{clip}. MaskCLIP+ \cite{zhou2021denseclip} adapts a frozen CLIP model and leverages pseudo-per-pixel labeling for semantic segmentation. Additionally, GroupViT \cite{groupvit} and OpenSeg \cite{openseg} learn segmentation masks from large-scale text supervision. In contrast to these approaches, we generate segments by only distilling the knowledge from CLIP vision encoder.

\noindent \textbf{Denoising autoencoder.} Denoising autoencoders~\cite{mae,lomar,beit,peco} have gained popularity as a means of reconstructing original images from corrupted inputs. This technique is widely used in representation learning. There are various denoising strategies including jigsaw puzzles~\cite{noroozi2016unsupervised}, inpainting~\cite{pathak2016context}, and color restoration~\cite{larsson2016learning}, etc. Among these strategies, MAE~\cite{mae}, or masked autoencoder, stands out for its ability to reconstruct missing patches with superior performance. MAE also improves training efficiency by reducing the number of input tokens in the encoder. Our ZeroSeg also builds upon the success of MAE and incorporates a masked autoencoder to improve the training efficiency and semantic representation for those segments.

%% file: Method.tex


\begin{figure*}[t!]
\centering
\begin{minipage}{0.67\textwidth}
\centering
\includegraphics[width=1\linewidth]
    {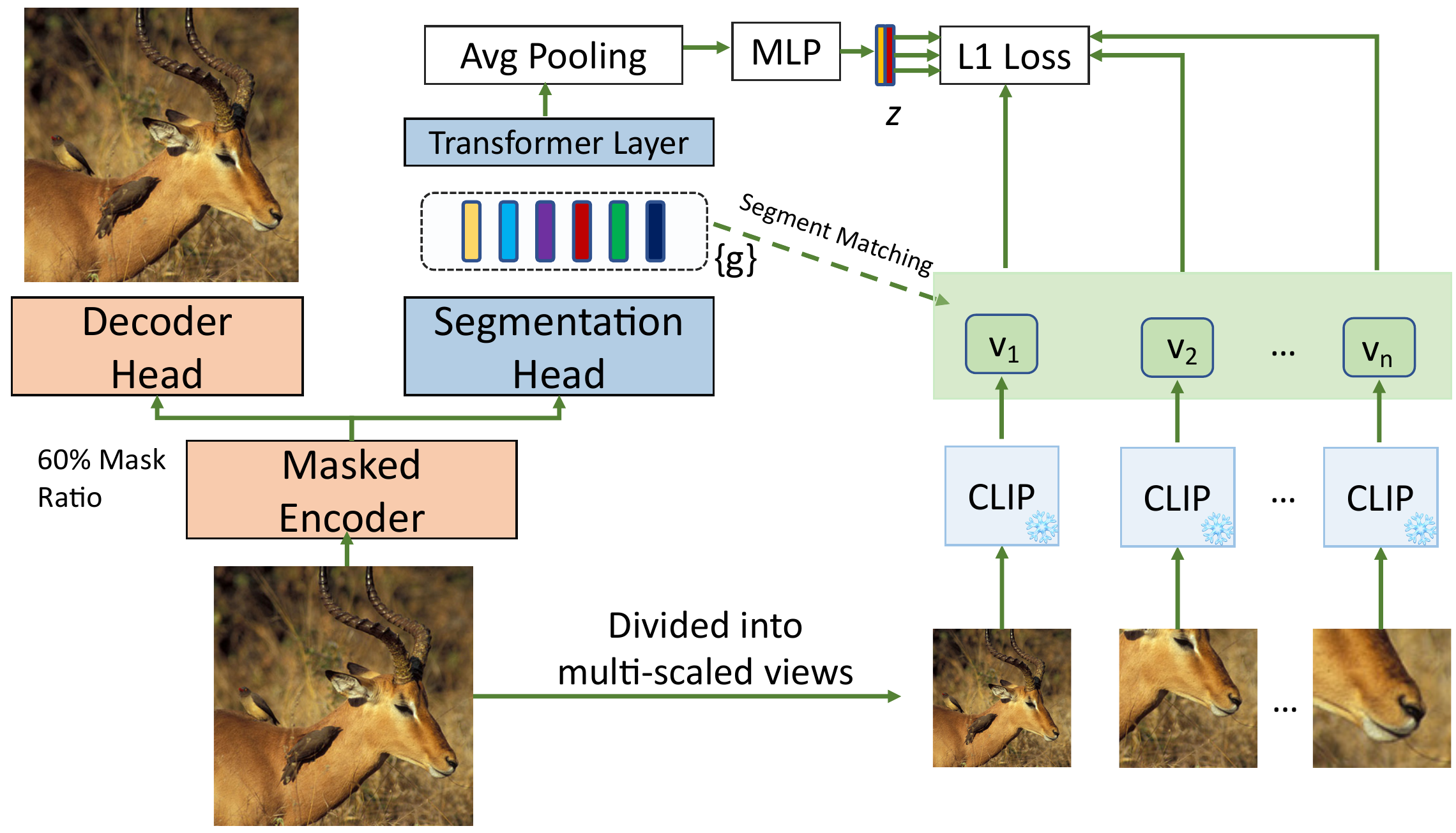} 
\end{minipage} \hfill
\begin{minipage}{0.3\textwidth}
\centering
 \includegraphics[width=0.9\linewidth]
{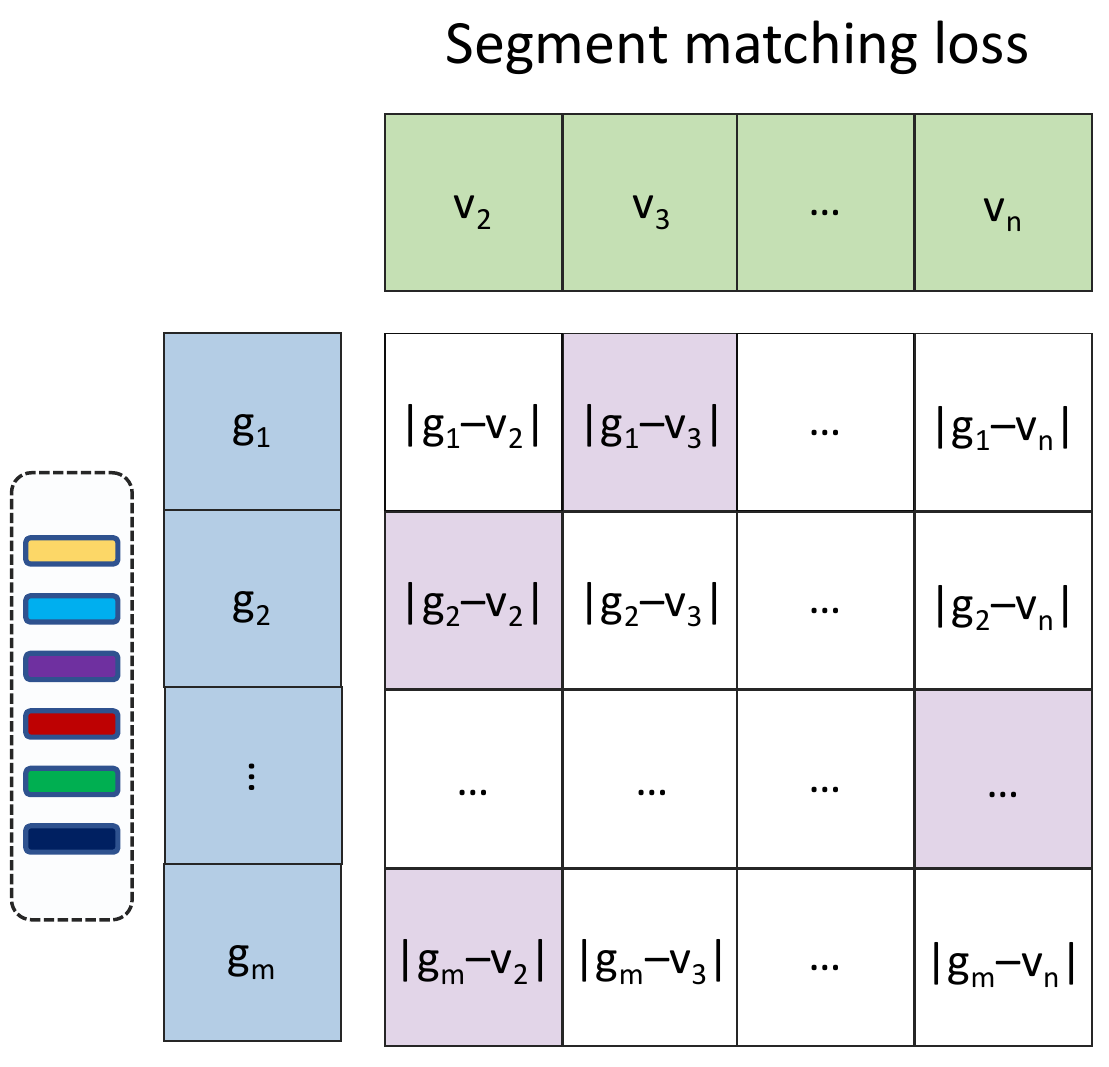}   
\end{minipage} \hfill
\vspace{1em}
\caption{\textbf{Training ZeroSeg model.}
ZeroSeg architecture consists of a ViT encoder and two heads including a decoder head and a segmentation head. 
The outputs from the decoder head is used to reconstruct the masked input image during training (\ie, masked autoencoding~\cite{mae}),
while the outputs from the segmentation head are transformed into several segment tokens $\{g\}$ to learn semantic segmentation via distillation. 
To effectively distill localized semantic information to the segmentation model, 
ZeroSeg employs a multi-scale feature generation method that divides the input image into multi-scale views, using \eg 2$\times$2 and 3$\times$3 grids,  
and pass these views to a pretrained CLIP visual encoder to produce visual features \{$v_1$, $v_2$, ... ,$v_n$\}. 
Then, ZeroSeg distills semantic information from these multi-scale features to the segmentation model via two loss functions.
The first one is an L$_1$ distillation loss between $\{v_1, v_2, v_3, ... , v_n\}$ and the global feature $z$. 
The second one is a segment matching loss to perform distillation between local region features $\{v_2, v_3, ... , v_n\}$ (excluding $v_{1}$ since it corresponds to the full-sized image feature) and segment tokens.
For each segment token, this loss function searches for its nearest neighbor local region, and minimizes the L$_1$ distance between them.
}
\vspace{-1em}
\label{main_arch}
\end{figure*}

This section presents our proposed architecture, ZeroSeg, which learns to perform semantic segmentation by only distilling the knowledge from the CLIP vision encoder. 
The architecture of ZeroSeg is illustrated in Figure \ref{main_arch}. 
ZeroSeg incorporates a masked encoder \cite{mae} as the main backbone, and it has two different heads, the first one is the reconstruction decoder for reconstructing the masked patches. 
The other one is the segmentation head to learn the semantic segmentation task. By incorporating the masked encoder-decoder, we empirically found that it can generate more reliable segmentations while being more efficient.
During training, only a fraction (40\%) of the visual patches are fed into the encoder, while the masked decoder reconstructs the remaining patches. 
We divide the full image into grids of multiple scales, and then compute images features from these grids. 
Next, we distill the grid features into the ZeroSeg model with mainly two losses. The first one is a multiscale feature distillation loss, while the other one is a segment matching loss to promote the semantic consistency between the segments and the visual concepts from the CLIP visual encoder. 


\subsection{Architecture}
We build our ZeroSeg model based upon the recent masked autoencoder (MAE) work~\cite{mae}, which aims to learn semantically meaningful representations through reconstructing masked-out image pixels. 
Similar to MAE, ZeroSeg leverages an asymmetric encoder-decoder architecture (Fig.~\ref{main_arch} left). When presented with an image, the encoder divides it into a sequence of non-overlapping patches. The encoder then selects a subset of visual tokens from each patch as input and generates the corresponding latent representation. Subsequently, the decoder utilizes this latent representation to reconstruct the missing patches, thereby producing a reconstructed image. 
ZeroSeg then trains the model by minimizing the mean squared error (MSE) between the reconstructed image and the original image in the pixel space, with the expectation that the resulting encoder would produce useful semantic representation that could benefit downstream tasks. 

In addition to the encoder-decoder structure tailored for mask autoencoding, we also incorporate an important segmentation head design (Fig.~\ref{main_arch} right) to help ZeroSeg learn to perform open-vocabulary semantic segmentation.

To group visual concepts, we build upon the previous work GroupViT~\cite{groupvit}.  
This approach involves organizing grouping layers into a hierarchy of stages, with each stage containing a grouping block to combine smaller groups into larger ones. 
Specifically, at each grouping layer, learnable segment tokens are used to bring semantically similar tokens together to form a single segment token based on their similarity.
Finally, the image segments are merged into a fixed number of segment tokens ${\{\text{g}_1, \text{g}_2,...,\text{g}_m\}}$, each corresponding to a disjoint image region. 
This grouping process enables the method to organize visual information into arbitrary semantically meaningful image segments. 

Though successful, GroupViT requires a large set of image-caption pairs for training, which is cumbersome and, as we will show, introduces bias into the type of data included in the training set that ultimately hurts the performance on the segmentation task. 
For this reason, we propose a text-free segmentation head in ZeroSeg (shown in Fig.~\ref{main_arch}). 
This means that all we need for training is a set of unlabeled images, which simplifies the training and makes our method much more widely applicable.   
Specifically, to derive the semantic representation for segment tokens, we extract multi-scale image features using a pretrained CLIP visual encoder and distill them into these tokens. 
Since CLIP visual encoders are trained to produce representations matching the text encoder outputs, we leverage this to produce the ``pseudo text supervision'' and thus avoid any text annotations.    

\subsection{Multi-scale image feature distillation} 
\noindent \textbf{Multi-scale image feature extraction.} 
An image can contain complex and diverse semantic information. 
Since the CLIP model only provides a single global representation for the entire image, 
it may not be sufficient to extract detailed regional semantic information. 
As we will show in experiments, it's inadequate to naively adapting the CLIP model to our context, as it fails to capture the concept specific (\ie, objects or stuff) information which is critical for semantic segmentation. 
To address this limitation, we propose a multi-scale image feature extraction strategy to better capture regional semantic information at different scales. 
Specifically, this strategy involves dividing the full image into multiple views, such as 2x2, 3x3 grids, each corresponds to a different sub-region of the full image, as illustrated in Fig.~\ref{main_arch}. 
We then resize each view into a full-size image, and pass them through the CLIP visual encoder to produce image features of different scales: ${\{\text{v}_1, \text{v}_2,...,\text{v}_n}\}$, 
which are more likely to capture diverse objects and extract more object-localized semantic information.

\begin{table*}[t!]
\small 
\begin{center}
\resizebox{0.95\textwidth}{!}{
\begin{tabular}{lccccccccc}
\toprule
\multicolumn{5}{c}{Pretraining} & & \multicolumn{4}{c}{Transfer Learning} \\
Models & Arch  & Dataset  & Scale  & Supervision & Require labels & Zeroshot &  VOC &  Context  & COCO \\
\midrule
 $\text{DeiT}^{\#}$ \cite{touvron2021training} &ViT & IN-1K \cite{imagenet} & 1.3M & class  & Yes & \xmark & 53.0 & 35.9& - \\
 $\text{DINO}^{\#}$ \cite{dino} & ViT  & IN-1K & 1.3M & self & Yes & \xmark & 39.1 & 20.4 & -  \\
 $\text{MoCo}^{\#}$ \cite{moco} & ViT &  IN-1K & 1.3M& self & Yes & \xmark & 34.3 & 21.3 & -\\
   \textcolor{Gray2}{MaskCLIP+} \cite{dong2022maskclip} & \textcolor{Gray2}{ViT} & \textcolor{Gray2}{Context+COCO+IN-22k} &\textcolor{Gray2}{14M} & \textcolor{Gray2}{pseudo masks} & \textcolor{Gray2}{Yes} & \textcolor{Gray2}{\xmark} &  \textcolor{Gray2}{-}& \textcolor{Gray2}{31.1} & \textcolor{Gray2}{18.0} \\
 \midrule

  GroupViT  & ViT &  CC12M+YFCC & 26M & text & Yes & \cmark & 52.3 & 22.4 &  24.3 \\
  \midrule
   CLIP & ViT & LAION-20M~\cite{dong2022maskclip} & 20M & text & Yes & \cmark & - &  13.5 & 8.2 \\
  MaskCLIP \cite{dong2022maskclip} &ViT & LAION-20M \cite{dong2022maskclip} &20M & text & Yes & \cmark &  - & 17.7 & 11.8 \\
 $\text{GroupViT}^{*}$ \cite{groupvit}  & ViT & CC3M+COCO & 3.4M & text & Yes &\cmark & 28.1 & 14.8 & 12.9\\
 SegCLIP \cite{segclip} & ViT & CC3M+COCO & 3.4M & text+$\text{CLIP}_{\text{T}}$ & Yes &  \cmark & 33.3 & 19.1 &15.2 \\
 \rowcolor{Gray}
ZeroSeg (Ours)  & ViT  & CC3M+COCO & 3.4M & $\text{CLIP}_{\text{V}}$ & \textcolor{Green}{No}& \cmark & 37.3& 19.7 & 17.8  \\
\rowcolor{Gray}
ZeroSeg (Ours)  & ViT  & IN-1K & 1.3M & $\text{CLIP}_{\text{V}}$ & \textcolor{Green}{No}& \cmark & \textbf{40.8} & \textbf{20.4} & \textbf{20.2}
\\

\bottomrule
\end{tabular}
}
\end{center}
\caption{
\textbf{Comparison to state-of-the-arts baselines.}
In the top section, we compare ZeroSeg to fully supervised segmentation methods. 
Whereas in the middle and bottom sections, 
we compare ZeroSeg to zero-shot segmentation methods which do not require any finetuning or adaption on target segmentation datasets.   
Note that MaskCLIP+ training requires a pretrained MaskCLIP model to generate pseudo segmentation ground truth and an adaption step on target segmentation datatsets.
$\text{CLIP}_{\text{V}}$ and $\text{CLIP}_{\text{T}}$ denote the visual and text encoder of a pretrained CLIP model, respectively.
$\#$ refers to numbers reported in GroupViT~\cite{groupvit}, while $*$ refers to results reported from SegCLIP~\cite{segclip}.
All results are reported using the mIoU metric.
} 
\label{main_results}
\end{table*}

\noindent \textbf{Multi-scale feature distillation loss.} 
To leverage the semantic information in the multi-scale CLIP visual features, 
we adopt a Transformer layer to encode all segment tokens, followed by an average pooling and an MLP layer to obtain the global image representation $z$. 
We then compute the multi-scale feature distillation loss between $z$ and the set of multi-scale image features ${\{\text{v}_1, \text{v}_2,...,\text{v}_n\}}$. 
For each $\text{v}$, we distill its knowledge to $z$ using an L$_1$ loss.
This process compels the global image feature $z$ to capture diverse and distinct regional semantic representations, 
thereby contributing to a more comprehensive semantic understanding of the image.

\noindent \textbf{Segment matching loss.} 
The current top-down approach for learning semantic masks with segment tokens lacks object-grounded constraints, 
which can potentially result in inconsistent semantic regions being captured by each segment token (\eg, mask pixels leaking into neighboring objects). 
This inconsistency can lead to incorrect segment classification. 
To overcome this, we propose a new segment matching loss $\mathcal{L}_{match}$ as follows:

\begin{equation}
    \label{equation}
    \mathcal{L}_{match} =  \sum_{i=1}^{m} \min_{j} \left| \text{g}_i - \text{v}_j \right|
\end{equation}

$\mathcal{L}_{match}$ aims to map each segment token $\text{g}_i$ to its most semantically aligned multi-scale image region feature $\text{v}_{j}$, as illustrated in Fig \ref{main_arch} (right). 
Note that this segment matching loss is only computed between each segment token $\text{g}_i$ and local-regional features excluding the full-size image features. This design is to encourage each segment token to capture more object-centric semantic information.
We achieve this by minimizing the L$_1$ distance between each $\text{g}_i$ and its nearest $\text{v}_j$, also measured in L$_1$ distance. 
As we will show in Sec.~\ref{sec:ablations}, adding this segment matching loss largely helped improve the semantic segmentation accuracy, 
by avoiding poor matches between segment tokens and image regions during training.  







%% file: Experiments.tex

\subsection{Implementation details}

\noindent \textbf{Model architecture.} 
Our proposed model, ZeroSeg, is based on the ViT-base architecture \cite{dosovitskiy2021an}. 
We use a 12-layer ViT transformer as our encoder. 
While for the reconstruction and segmentation heads, we adopt two transformer decoders each consisting of 8 and 5 transformer layers, respectively. 
Two grouping stages are appended to the segmentation head after the 2nd and 4th transformer layers, employing 32 and 8 learnable group tokens, respectively. 
To encode the positional information of image patches, 
we utilize absolute positional encoding~\cite{vaswani2017attention} for both the encoder and the masked decoder. 
Multi-scale image features are extracted using a pretrained CLIP-L vision encoder. Details on the specific hyperparameters can be found in our Supplementary Materials.

\noindent \textbf{Training details.} 
In our work, we mainly train our ZeroSeg model on images from ImageNet 1k \cite{imagenet} dataset. 
We also train on CC3M \cite{conceptual} and COCO \cite{coco} for ablation study. 
We train our model on ImageNet-1K dataset for 80 epochs, with the first 20 epochs as the warm-up period, during which we use a base learning rate of 1.5e-4. 
We use the AdamW optimizer and a batch size of 4096. 
We only employ the center crop without any other augmentation strategies, 
hence we can pre-compute and cache the multi-scale image features using the CLIP model for better training efficiency. 
Finally, all training images are rescaled to 224$\times$224 during training.


\subsection{Comparison to the state-of-the-arts}
We evaluate ZeroSeg on three benchmark datasets: PASCAL VOC 2012 \cite{voc}, PASCAL Context \cite{context}, and COCO \cite{coco}. 
These datasets consist of 20, 59, and 80 foreground classes, respectively. 
To generate text embeddings for each class $c$ during inference, 
we feed the classes to the CLIP text encoder using a set of predefined prompt templates (\eg, ``a photo of the \{class\}'') and produce the corresponding class embeddings $t_c$, $c \in \{1, 2, ..., C\}$, 
where $C$ is the total number of foreground classes. 
We then compute the cosine similarity between each group token $\text{g}_m$ and class embedding $t_c$. 
Following~\cite{groupvit}, we adopt a threshold to filter out the background class and then take the nearest neighbor class as the semantic label for each group token. 
Specifically, we set the threshold to 0.95 for PASCAL VOC, 0.05 for PASCAL Context and 0.35 for COCO.  
All images are resized to have a shorter side length of 448 during inference. 

We compare our ZeroSeg model to various supervised and weakly-supervised semantic segmentation methods, 
including DeiT \cite{touvron2021training}, DINO \cite{dino}, MoCo \cite{moco}, GroupViT \cite{groupvit}, MaskCLIP \cite{dong2022maskclip}, MaskCLIP+ \cite{zhou2021denseclip} and SegCLIP \cite{segclip}. 
Notably, our ZeroSeg model is the only one method that does not require any form of human labels during the training process. 
For fair comparisons, all models are using the same ViT architecture as the backbone~\cite{dosovitskiy2021an}. 

Table~\ref{main_results} summarizes the results of our comparison. 
First, the results demonstrate that ZeroSeg can achieve competitive performance to several non-zero-shot supervised segmentation baselines, despite not using any segmentation label during training. 
Specifically, ZeroSeg achieved an mIoU score of 40.8 on VOC, surpassing the performance of the supervised segmentation model with DINO and MoCo pretraining by +1.7 and +6.5 mIoU, respectively. 
Comparing to other zero-shot segmentation methods, ZeroSeg outperforms all baselines with a large margin when trained on a similar amount of data. 
For example, when trained on CC3M+COCO, ZeroSeg outperforms GroupViT and SegCLIP on VOC by +9.2\% and +4.0\%, respectively. 
In fact, ZeroSeg even outperforms MaskCLIP (+2.7 on PASCAL Context, +8.4 on COCO) which is trained on 15$\times$ more data (1.3M \vs 20M). 
These results demonstrate that our ZeroSeg model not only learns strong zero-shot segmentation capability, but also achieves so with high data efficiency. 
Finally, an interesting observation is that training on 1.3M ImageNet images yield better results compared to training on 3.4M images from CC3M and COCO, 
we hypothesize that this is due to the fact that ImageNet contains more common objects compared to Conceptual Caption, 
making it more aligned to objects seen in popular semantic segmentation benchmarks. 
This also highlights the advantage of not relying on texts during training, as it allows ZeroSeg to be trained on the widest possible range of data sources.



\begin{table}[t!]
\begin{center}
\begin{tabular}{lcc}
  Model & GroupViT & ZeroSeg (ours) \\
  \toprule
  \#votes & 323/1000  & 677/1000 \\
\end{tabular}
\end{center}
\caption{
\textbf{Human study for open-vocabulary segmentation.} 
We compare the number of favoring votes received by ZeroSeg and GroupViT,
when asking AMT workers to evaluate the quality of segmentation results on sampled images from Conceptual Caption.
} 
\label{human_evaluation}
\end{table}

\begin{table}[t!]
\begin{center}
\begin{tabular}{lc}
Window Scales  & VOC (mIoU) \\
\toprule
1x1 & 21.1 \\
1x1+2x2 & 23.7 \\
1x1+3x3 & 40.2 \\
1x1+4x4 & 32.4 \\
\rowcolor{Gray}
1x1+2x2+3x3+4x4 & \textbf{40.8}\\ 
\end{tabular}
\end{center}
\caption{
\textbf{Ablating multi-scale image features.} 
We dissect the impact of different settings to compute the multi-scale image feature.  
As an example, 2$\times$2 refers to the setting where the full image is divided into 2$\times$2 non-overlapping grids. 
Note that the segment matching loss is applied to all settings except for the 1$\times$1 grid.
} 
\vspace{-2em}
\label{multi_scale_ablation}
\end{table}


\subsection{Open-vocabulary semantic segmentation}\label{sec:openvocabseg}
Due to the high annotation costs, popular semantic segmentation datasets all have relatively small vocabulary (\eg, 20 and 59 classes for PASCAL VOC and Context). 
This means that it still remains relatively unexplored on how segmentation models perform in an open vocabulary setting.   
Though as an important task with great practical values, it's non-trivial to conduct evaluation for open-vocabulary semantic segmentation. 
Therefore, to facilitate the evaluation, we simulate the open-vocabulary setting by constructing a large vocabulary consisting of 1000 classes from ImageNet~\cite{imagenet}, 
and compare ZeroSeg against the GroupViT baseline using this vocabulary. 
For test images, we randomly sample 200 images from the Conceptual Caption validation set.
We generate segmentation masks using both our ZeroSeg model trained on 1.3M ImageNet images, 
and the GroupViT model trained on 26M image-text pairs from CC12M \cite{conceptual}+YFCC \cite{yfcc}. 
Since there are no ground-truth segmentation labels for Conceptual Caption, 
we conduct a human study to evaluate the quality of the generated segmentations. 
Specifically, we resort to Amazon Mechanical Turk for this. 
We assign each image with overlaid segmentation masks to 5 different workers, and ask each worker to decide which one in the pair has better segmentation quality. 

Table~\ref{human_evaluation} displays the evaluation results of our study. The results demonstrate that ZeroSeg received a larger number of votes than GroupViT (68\% \vs 32\%), indicating that ZeroSeg is capable of generating more reliable and human-preferable segmentation, particularly when dealing with a large vocabulary. These findings highlight the open-vocabulary benefits of transferring knowledge from large-pretrained vision-language models.



\subsection{Ablation study}\label{sec:ablations}


\begin{table}[t]
\begin{center}
\begin{tabular}{lc}

Ablations  & VOC (mIoU) \\
\toprule
Base & 21.1 \\
Base+Multi-scale  & 28.5  \\
Base+segment matching  &  38.6 \\
\rowcolor{Gray}
Base+Multi-scale+segment matching & \textbf{40.8} \\
\end{tabular}
\end{center}
\caption{
\textbf{Ablating distillation losses}. 
`Base' refers to the setting where distillation is applied only between the full image feature and the global image representation $z$. 
Meanwhile, `Multi-scale' refers to that the distillation is applied between all multi-scale features and the global representation $z$. 
Finally, `segment matching' refers to turning on the segment matching loss computed between each segment token and the multi-scale image features.
} 
\vspace{-1em}
\label{nearest_neighbor_ablation}
\end{table}

\noindent \textbf{Impact of multi-scale image feature distillation.}
In this study, we explore the impact of different designs for the multi-scale image feature distillation method. 
Specifically, we vary the number and the size of the grids used to compute the multi-scale features. 
For example, ``1$\times$1+2$\times$2'' refers to combining the full image feature (1$\times$1) and features computed from each of the 2$\times$2 grids.  
All ablative results are presented in Table~\ref{multi_scale_ablation}. 
Our finding suggests that it's insufficient to produce accurate semantic segmentation, 
when we only distill the knowledge to our ZeroSeg model from a full-sized image feature (1$\times$1), 
as it fails to capture enough localized semantic features. 
Therefore, we explore more grid size settings such as 2$\times$2, 3$\times$3, and 4$\times$4, 
as they are supposed to capture different levels of object details in the image. 
When combined with the full image feature (1$\times$1), we observe that 3$\times$3 grids outperform other settings (40.2 mIoU),
while it works the best when we combine all grid sizes to produce multi-scale features for distillation.
Overall, Table~\ref{multi_scale_ablation} demonstrates that the multi-scale image feature design has a significant impact on the success of distillation, 
as it almost doubled the segmentation mIoU on VOC (21.1 to 40.8). 


\noindent \textbf{Impact of segment matching loss.}
We compare the performance of our model with and without the segment matching loss. 
The results are presented in Table~\ref{nearest_neighbor_ablation}. 
We first compare the \textit{base} to the \textit{base + multi-scale} setting. 
\textit{base} refers to the setting in which we only distill knowledge from the full image feature (\ie, 1$\times$1 grid) to the global image representation $z$ (Fig.~\ref{main_arch}). 
Whereas \textit{multi-scale} refers to the distillation loss between the multi-scale image features (2x2, 3x3 and 4x4 grid features) and $z$.
Our findings indicate that including the segment matching loss results in a large improvement over the model's performance. Specifically, the addition of the segment matching loss led to a 17.5 mIoU increase on PASCAL VOC over the \textit{base} model. Additionally, the segment matching loss also improves the performance of the \textit{base  + multi-scale} setting by 12.3 mIoU. These results suggest that the segment matching loss plays a crucial role in effectively transferring visual concept knowledge from CLIP to segment tokens. Overall, this ablative experimental result highlights the importance of the segment matching loss for our model's success.


\begin{table}[t]
\small
\begin{center}
\begin{tabular}{l@{\hspace{0.7em}}c@{\hspace{0.7em}}c@{\hspace{0.7em}}c@{\hspace{0.7em}}c@{\hspace{0.7em}}c@{\hspace{0.7em}}c@{\hspace{0.7em}}c@{\hspace{0.7em}}c}
Mask ratio & 0\% & 20\% & 30\% & 40\% & 50\% & 60\% & 70\% & 80\%  \\
\toprule
mIoU & 35.6 & 37.6 & 38.7 & 40.4 & 39.8 & \textbf{40.8} & 33.3 & 32.8\\
Speedup (\%) & 0  & 15 &19 & 26& 32 & 36 & 39 & 43 \\
\end{tabular}
\end{center}
\caption{
\textbf{Ablating mask ratios}. 
We study the impact of different mask ratios on segmentation quality (mIoU on VOC~\cite{voc}) and the training speed. 
The relative speed-up is measured on the full model,
by comparing to the setting of mask ratio being 0.
}
\label{mask_ratio}
\vspace{-1.5em}
\end{table}


\noindent \textbf{Mask ratio for encoder input.} 
As shown in~\cite{mae}, the mask ratio for the encoder input plays an important role affecting both the representation quality and the efficiency. 
We ablate the impact of different mask ratios on semantic segmentation accuracy in Table~\ref{mask_ratio}. 
The results suggest that a mask ratio of 60\% leads to the best accuracy at an mIoU of 40.8\% on VOC, while providing a $\sim$36\% speedup compared to the training  without any masks, and it also brings a improvement over 5.2 mIoU. Therefore, we choose 60\% mask ratio as our default mask ratio for all the future experiments.
Note that this is lower than the 75\% mask ratio used in the MAE paper~\cite{mae}, 
suggesting that it requires seeing more pixels (\ie, lower mask ratio) to better learn pixel-level tasks.








\subsection{Qualitative visualizations}
\begin{figure*}[t]
\begin{minipage}{0.5\textwidth}
\centering
 \includegraphics[width=1.0\linewidth]
    {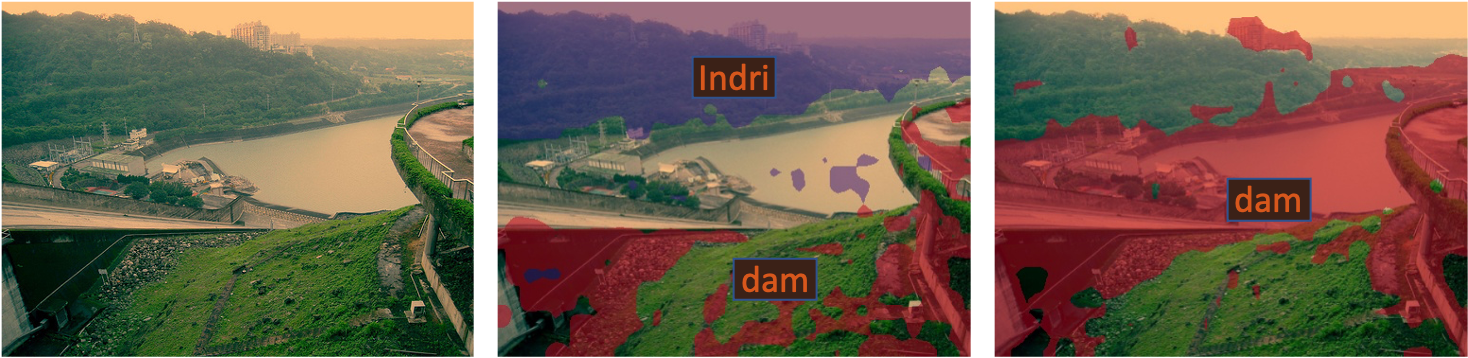}   
\end{minipage} \hfill
\begin{minipage}{0.5\textwidth}
\centering
 \includegraphics[width=1.0\linewidth]
    {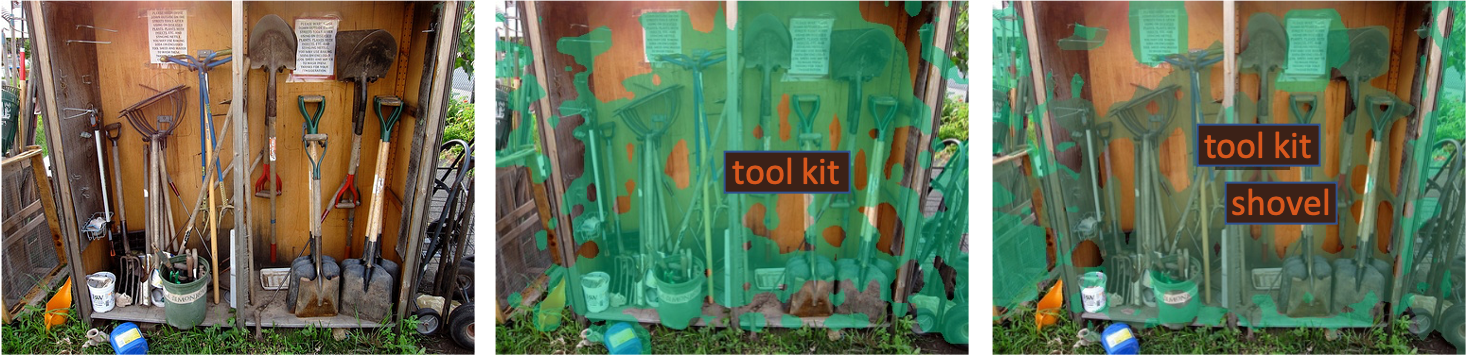}   
\end{minipage} \hfill
\begin{minipage}{0.5\textwidth}
\centering
 \includegraphics[width=1.0\linewidth]
    {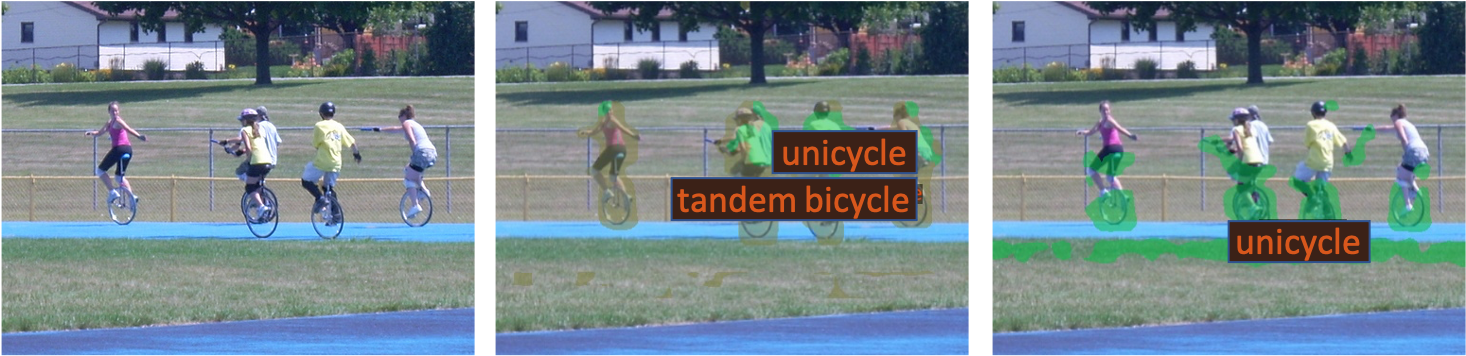} 
\end{minipage} \hfill
\begin{minipage}{0.5\textwidth}
\centering
 \includegraphics[width=1.0\linewidth]
    {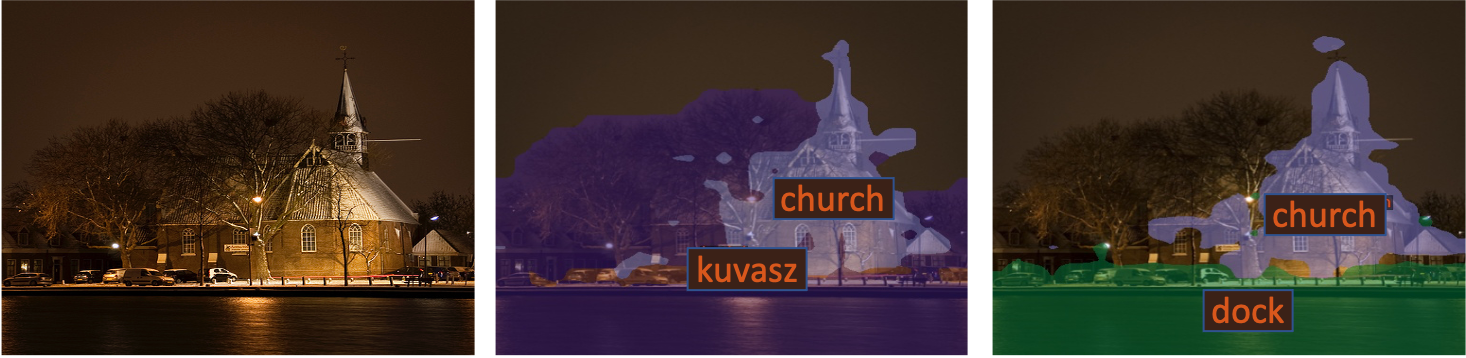}   
\end{minipage} \hfill

\begin{minipage}{0.5\textwidth}
\centering
 \includegraphics[width=1.0\linewidth]
    {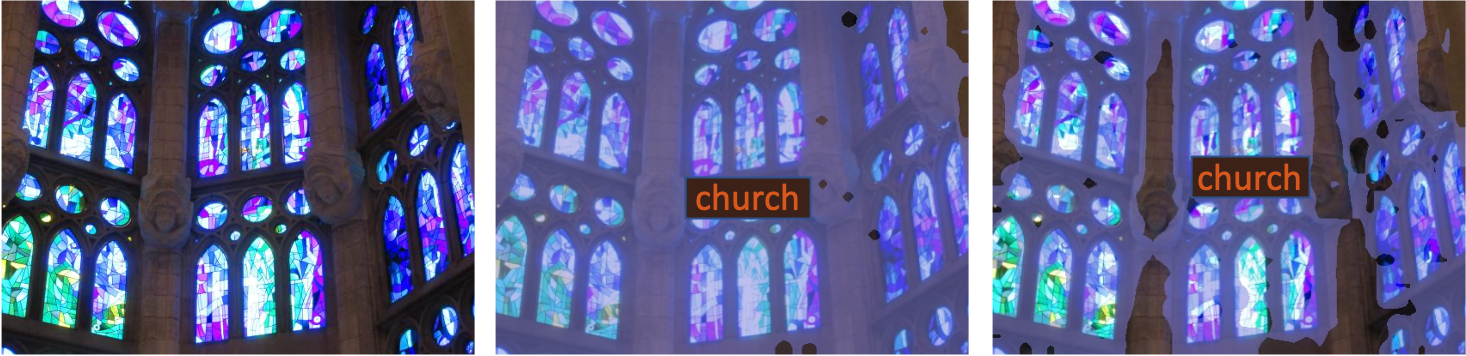}   
\end{minipage} \hfill
\begin{minipage}{0.5\textwidth}
\centering
 \includegraphics[width=1.0\linewidth]
    {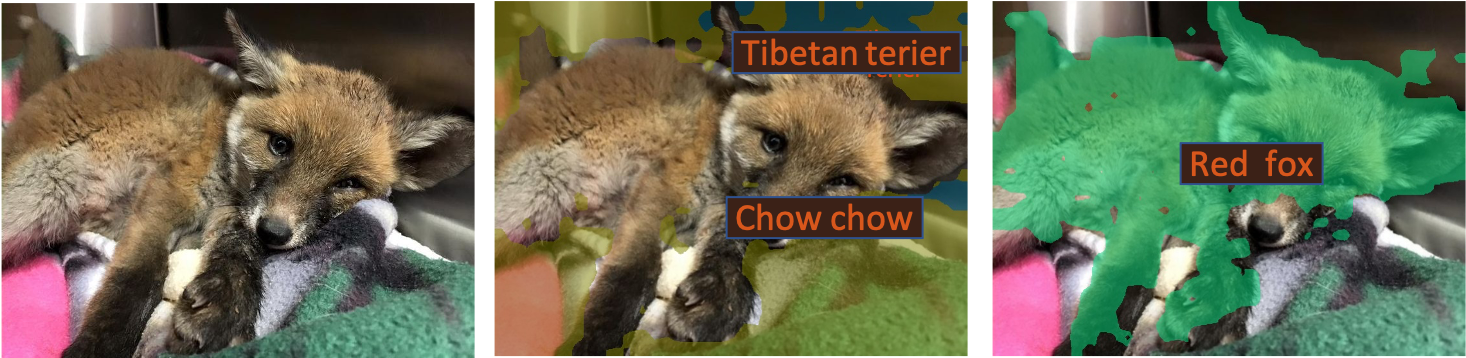}   
\end{minipage} \hfill
\begin{minipage}{0.5\textwidth}
\centering
 \includegraphics[width=1.0\linewidth]
    {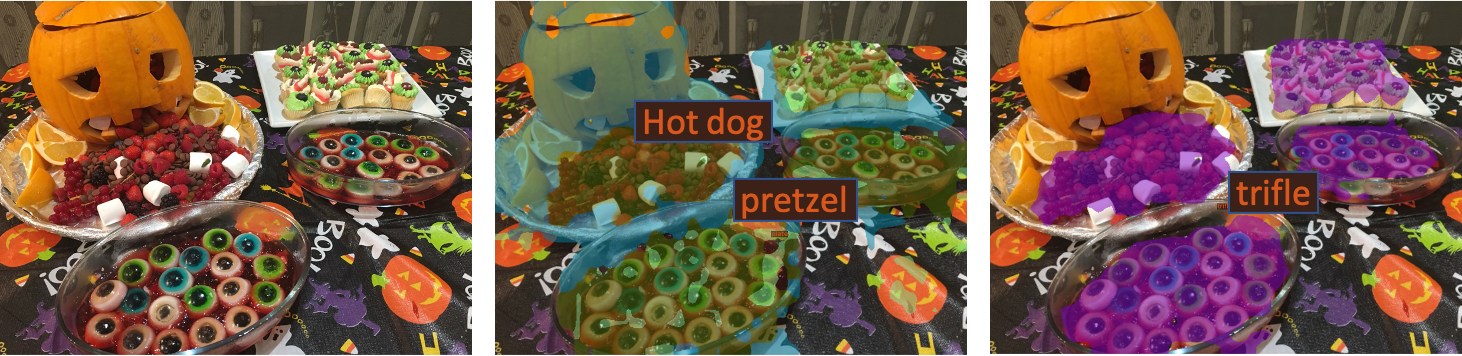} 
    \title{ Input  \hspace{40pt}  GroupViT  \hspace{40pt} ZeroSeg}
\end{minipage} \hfill
\begin{minipage}{0.5\textwidth}
\centering
 \includegraphics[width=1.0\linewidth]
    {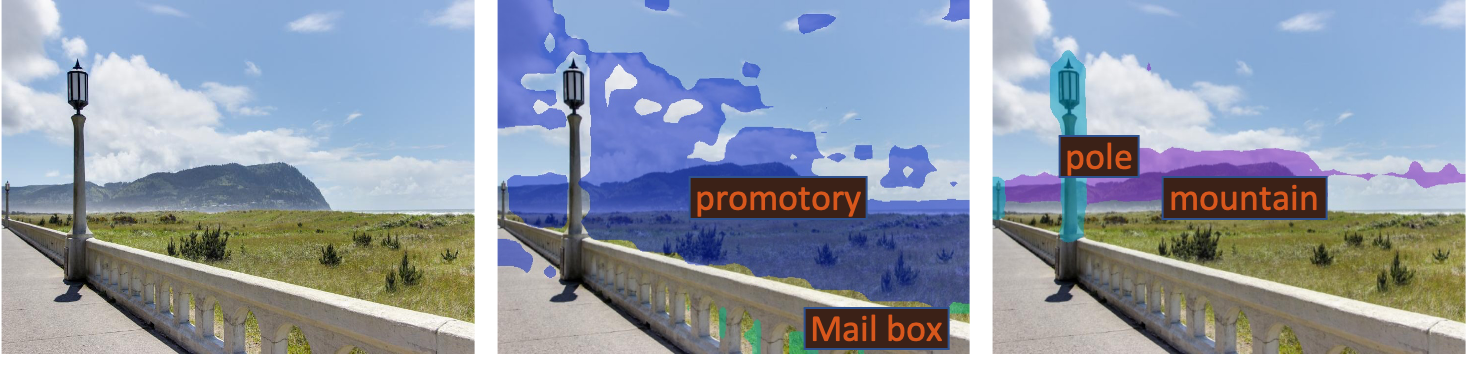}   
    \title{Input  \hspace{40pt}  GroupViT  \hspace{40pt} ZeroSeg}
\end{minipage} \hfill
\caption{
\textbf{Comparing GroupViT and ZeroSeg on open-vocabulary semantic segmentation.} 
We present a comparison on open-vocabulary semantic segmentation between GroupViT and our ZeroSeg model.  
To simulate the open-vocabulary setting, we use a large vocabulary comprising 1000 classes from ImageNet. 
Half of the test images are sampled from the ImageNet validation set (top 2 rows), while other half from the Conceptual Caption dataset (bottom 2 rows).
For each image, we show the original input, the output from GroupViT, and the output from our ZeroSeg model. 
}
\label{large_vocabulary}
\vspace{-1.5em}
\end{figure*}

\noindent \textbf{Visualizing open-vocabulary semantic segmentation.}
In addition to the human study results described in Sec.~\ref{sec:openvocabseg}, 
we present qualitative visualizations for open-vocabulary segmentation in this section.  
To do this, we apply both our ZeroSeg model and the GroupViT model (using the publicly released weights) to perform zero-shot open-vocabulary semantic segmentation. 
In Fig. \ref{large_vocabulary}, we visualize the results on 4 images randomly sampled from both the ImageNet and the Conceptual Caption validation set. 
From the figure, it's clear that ZeroSeg produces better results under the open-vocabulary setting, 
as it inherited CLIP model's extraordinary capability for fine-grained classification. 
For example, in the top-right image, 
ZeroSeg accurately predicts the \textit{shovel} class, rather than simply categorizing everything as \textit{toolkit}, which is the case for GroupViT.


\begin{figure}[t]
\centering
 \includegraphics[width=1.0\linewidth]
    {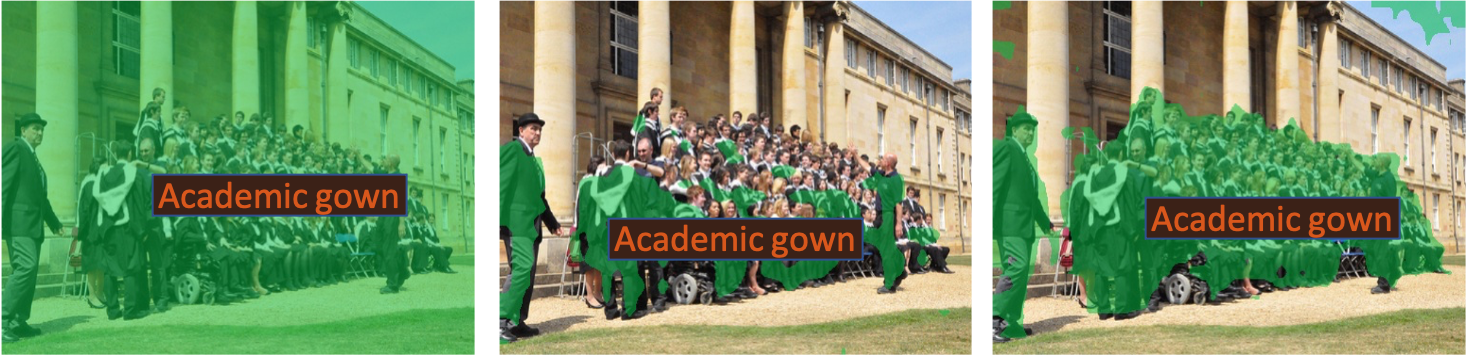}   
\centering
 \includegraphics[width=1.0\linewidth]
    {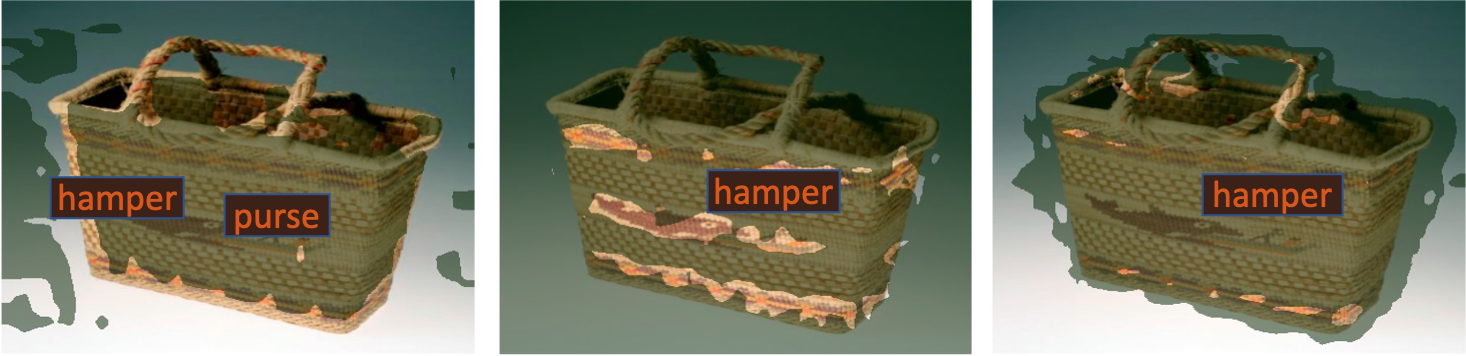}
\centering
 \includegraphics[width=1.0\linewidth]
    {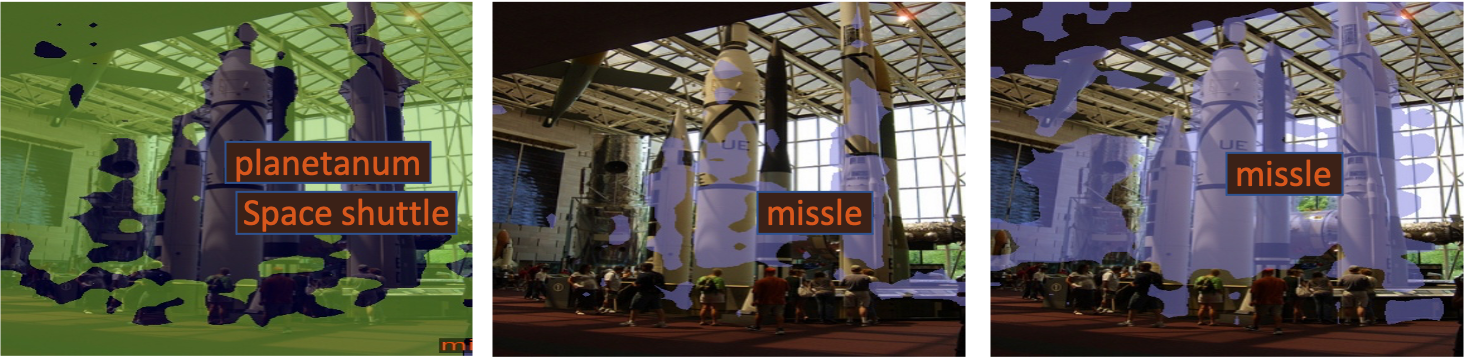}

\caption{
\textbf{Segmentation quality with different losses.} 
We present the qualitatively segmentation results from models trained with different loss functions.
Specifically, we compare models trained with only the global distillation loss (`Base'), 
adding in the multi-scale loss (`Multi-scaled'), 
and with the combined multi-scale and segment matching losses (`Multi-scaled + Segment matching'). 
}
\label{ablation_visualization}
\vspace{-1.5em}
\end{figure}



\noindent \textbf{Visualization of the ablation on loss functions.} 
To qualitatively observe the impact of different loss functions, 
we visualize the segmentation masks on two images, selected from the ImageNet validation set, using variants of ZeroSeg models trained with different loss functions. 
The visualizations are presented in Fig.~\ref{ablation_visualization}. 
We observe that the \textit{base} model is not able to produce meaningful segments, despite producing the correct object class label. 
With the \textit{multi-scale} loss added, the model starts to produce localized segments, but still lags behind on the precise delineation of object boundaries. 
Finally, by integrating both the \textit{multi-scale} and the \textit{segment matching} loss, 
our ZeroSeg model now produces much more accurate object boundaries, demonstrating the effectiveness of both losses. 





%% file: Discussion.tex
\begin{table}[t]
\small
\setlength{\tabcolsep}{3pt}
\begin{center}
\begin{tabular}{lccccc}
 mIoU &  bedclothes & ground & keyboard & motorbike & avg \\
\toprule
GroupViT & 0.91 & 9.33 & 7.39 & 21.47 & 9.78 \\
ZeroSeg  & \textbf{11.21} & \textbf{23.31} & \textbf{29.1} & \textbf{47.77} & 
\textbf{27.85} \\
\end{tabular}
\end{center}
\label{context}
\caption{Performance on semantic classes with sub-word or compound word.} 
\vspace{-2em}
\end{table}

In this work, we present ZeroSeg as a novel method for training open-vocabulary zero-shot semantic segmentation models without using any human labels. 
ZeroSeg learns to perform semantic segmentation by distilling knowledge from a large-scale pretrained vision-language model.
This is a challenging task since these VL models are usually trained at an image-level and are not designed for pixel-level tasks like semantic segmentation. 

To effectively distill visual knowledge from the pretrained VL model to our ZeroSeg model, 
we designed two loss functions: the multi-scaled feature distillation loss and the segment matching loss. 
The multi-scaled feature distillation loss helps ZeroSeg capture object-localized semantic information at different scales. 
On the other hand, the segment matching loss aims to help align each segment token with the corresponding image region, 
and thus produce spatially consistent semantic features.  
Through our experiments, we demonstrated that both losses are critical to achieving good segmentation accuracy and they are supplementary to each other. 

We train ZeroSeg on 1.3M ImageNet images and observe that it achieves comparable or better results, 
compared to those models that are either pretrained on much larger image-text pair datasets, 
or finetuned with segmentation labels in a supervised manner. 
Furthermore, through human study and visualizations, we demonstrate that ZeroSeg outperforms GroupViT on the task of open-vocabulary segmentation.

We also discovered that GroupViT struggles with object
classes that are defined by sub-words such as ‘ground’, a sub-word of background, or compound words like ‘bedclothes’, ‘keyboard’, and ‘motorbike’ in Table \ref{context}, which might stem from the misunderstanding in the language context during the model training. In contrast, ZeroSeg performs much better (+18.07\%) on those sub-word or compound words since training ZeroSeg does not rely on textual context.

Overall, with ZeroSeg, we demonstrated that it's possible to effectively train semantic segmentation models by transferring the knowledge from a pretrained, general-purpose vision-language model. 
We hope this could open a new door to leverage the recent trendy efforts on foundation models~\cite{bommasani-foundation2021} to benefit pixel-level downstream tasks like semantic segmentation. 


\noindent \textbf{Broader impact.}
Our model has the unique capability to learn segmentation from images without human labels, 
thus enabling use cases across diverse domains. 
However, it's important to acknowledge that the large pretrained vision-language models on which our model is based may perpetuate biases present in the training data. 
Therefore, mitigations like careful training data filtering is crucial to ensure the ethical use of our model.

\noindent \textbf{Acknowledgements.} We would like to express our sincere appreciation to Xinlei Chen and Saining Xie for providing their thoughtful suggestions.

%% file: egpaper_for_review.bbl
\begin{thebibliography}{10}\itemsep=-1pt

\bibitem{flamingo}
Jean-Baptiste Alayrac, Jeff Donahue, Pauline Luc, Antoine Miech, Iain Barr,
  Yana Hasson, Karel Lenc, Arthur Mensch, Katherine Millican, Malcolm Reynolds,
  Roman Ring, Eliza Rutherford, Serkan Cabi, Tengda Han, Zhitao Gong, Sina
  Samangooei, Marianne Monteiro, Jacob Menick, Sebastian Borgeaud, Andrew
  Brock, Aida Nematzadeh, Sahand Sharifzadeh, Mikolaj Binkowski, Ricardo
  Barreira, Oriol Vinyals, Andrew Zisserman, and Karen Simonyan.
\newblock Flamingo: a visual language model for few-shot learning.
\newblock In Alice~H. Oh, Alekh Agarwal, Danielle Belgrave, and Kyunghyun Cho,
  editors, {\em Advances in Neural Information Processing Systems}, 2022.

\bibitem{baek2021exploiting}
Donghyeon Baek, Youngmin Oh, and Bumsub Ham.
\newblock Exploiting a joint embedding space for generalized zero-shot semantic
  segmentation.
\newblock In {\em Proceedings of the IEEE/CVF international conference on
  computer vision}, pages 9536--9545, 2021.

\bibitem{beit}
Hangbo Bao, Li Dong, Songhao Piao, and Furu Wei.
\newblock {BE}it: {BERT} pre-training of image transformers.
\newblock In {\em International Conference on Learning Representations}, 2022.

\bibitem{bommasani-foundation2021}
Rishi Bommasani, Drew~A Hudson, Ehsan Adeli, Russ Altman, Simran Arora, Sydney
  von Arx, Michael~S Bernstein, Jeannette Bohg, Antoine Bosselut, Emma
  Brunskill, et~al.
\newblock On the opportunities and risks of foundation models.
\newblock {\em arXiv preprint arXiv:2108.07258}, 2021.

\bibitem{bucher2019zero}
Maxime Bucher, Tuan-Hung Vu, Matthieu Cord, and Patrick P{\'e}rez.
\newblock Zero-shot semantic segmentation.
\newblock {\em Advances in Neural Information Processing Systems}, 32, 2019.

\bibitem{dino}
Mathilde Caron, Hugo Touvron, Ishan Misra, Herv{\'e} J{\'e}gou, Julien Mairal,
  Piotr Bojanowski, and Armand Joulin.
\newblock Emerging properties in self-supervised vision transformers.
\newblock In {\em Proceedings of the IEEE/CVF international conference on
  computer vision}, pages 9650--9660, 2021.

\bibitem{conceptual}
Soravit Changpinyo, Piyush Sharma, Nan Ding, and Radu Soricut.
\newblock Conceptual 12m: Pushing web-scale image-text pre-training to
  recognize long-tail visual concepts.
\newblock In {\em Proceedings of the IEEE/CVF Conference on Computer Vision and
  Pattern Recognition}, pages 3558--3568, 2021.

\bibitem{chen2022visualgpt}
Jun Chen, Han Guo, Kai Yi, Boyang Li, and Mohamed Elhoseiny.
\newblock Visualgpt: Data-efficient adaptation of pretrained language models
  for image captioning.
\newblock In {\em Proceedings of the IEEE/CVF Conference on Computer Vision and
  Pattern Recognition}, pages 18030--18040, 2022.

\bibitem{lomar}
Jun Chen, Ming Hu, Boyang Li, and Mohamed Elhoseiny.
\newblock Efficient self-supervised vision pretraining with local masked
  reconstruction.
\newblock {\em arXiv preprint arXiv:2206.00790}, 2022.

\bibitem{chen2017rethinking}
Liang-Chieh Chen, George Papandreou, Florian Schroff, and Hartwig Adam.
\newblock Rethinking atrous convolution for semantic image segmentation.
\newblock {\em arXiv preprint arXiv:1706.05587}, 2017.

\bibitem{cheng2022masked}
Bowen Cheng, Ishan Misra, Alexander~G Schwing, Alexander Kirillov, and Rohit
  Girdhar.
\newblock Masked-attention mask transformer for universal image segmentation.
\newblock In {\em Proceedings of the IEEE/CVF Conference on Computer Vision and
  Pattern Recognition}, pages 1290--1299, 2022.

\bibitem{cheng2021per}
Bowen Cheng, Alex Schwing, and Alexander Kirillov.
\newblock Per-pixel classification is not all you need for semantic
  segmentation.
\newblock {\em Advances in Neural Information Processing Systems},
  34:17864--17875, 2021.

\bibitem{imagenet}
Jia Deng, Wei Dong, Richard Socher, Li-Jia Li, Kai Li, and Li Fei-Fei.
\newblock Imagenet: A large-scale hierarchical image database.
\newblock In {\em 2009 IEEE conference on computer vision and pattern
  recognition}, pages 248--255. Ieee, 2009.

\bibitem{dong2018few}
Nanqing Dong and Eric~P Xing.
\newblock Few-shot semantic segmentation with prototype learning.
\newblock In {\em BMVC}, volume~3, 2018.

\bibitem{peco}
Xiaoyi Dong, Jianmin Bao, Ting Zhang, Dongdong Chen, Weiming Zhang, Lu Yuan,
  Dong Chen, Fang Wen, and Nenghai Yu.
\newblock Peco: Perceptual codebook for bert pre-training of vision
  transformers.
\newblock {\em arXiv preprint arXiv:2111.12710}, 2021.

\bibitem{dong2022maskclip}
Xiaoyi Dong, Yinglin Zheng, Jianmin Bao, Ting Zhang, Dongdong Chen, Hao Yang,
  Ming Zeng, Weiming Zhang, Lu Yuan, Dong Chen, et~al.
\newblock Maskclip: Masked self-distillation advances contrastive
  language-image pretraining.
\newblock {\em arXiv preprint arXiv:2208.12262}, 2022.

\bibitem{dosovitskiy2021an}
Alexey Dosovitskiy, Lucas Beyer, Alexander Kolesnikov, Dirk Weissenborn,
  Xiaohua Zhai, Thomas Unterthiner, Mostafa Dehghani, Matthias Minderer, Georg
  Heigold, Sylvain Gelly, Jakob Uszkoreit, and Neil Houlsby.
\newblock An image is worth 16x16 words: Transformers for image recognition at
  scale.
\newblock In {\em International Conference on Learning Representations}, 2021.

\bibitem{voc}
Mark Everingham, Luc Van~Gool, Christopher~KI Williams, John Winn, and Andrew
  Zisserman.
\newblock The pascal visual object classes (voc) challenge.
\newblock {\em International journal of computer vision}, 88:303--308, 2009.

\bibitem{fu2019dual}
Jun Fu, Jing Liu, Haijie Tian, Yong Li, Yongjun Bao, Zhiwei Fang, and Hanqing
  Lu.
\newblock Dual attention network for scene segmentation.
\newblock In {\em Proceedings of the IEEE/CVF conference on computer vision and
  pattern recognition}, pages 3146--3154, 2019.

\bibitem{openseg}
Golnaz Ghiasi, Xiuye Gu, Yin Cui, and Tsung-Yi Lin.
\newblock Scaling open-vocabulary image segmentation with image-level labels.
\newblock In {\em Computer Vision--ECCV 2022: 17th European Conference, Tel
  Aviv, Israel, October 23--27, 2022, Proceedings, Part XXXVI}, pages 540--557.
  Springer, 2022.

\bibitem{mae}
Kaiming He, Xinlei Chen, Saining Xie, Yanghao Li, Piotr Doll{\'a}r, and Ross
  Girshick.
\newblock Masked autoencoders are scalable vision learners.
\newblock In {\em Proceedings of the IEEE/CVF Conference on Computer Vision and
  Pattern Recognition}, pages 16000--16009, 2022.

\bibitem{moco}
Kaiming He, Haoqi Fan, Yuxin Wu, Saining Xie, and Ross Girshick.
\newblock Momentum contrast for unsupervised visual representation learning.
\newblock In {\em Proceedings of the IEEE/CVF conference on computer vision and
  pattern recognition}, pages 9729--9738, 2020.

\bibitem{hu2020uncertainty}
Ping Hu, Stan Sclaroff, and Kate Saenko.
\newblock Uncertainty-aware learning for zero-shot semantic segmentation.
\newblock {\em Advances in Neural Information Processing Systems},
  33:21713--21724, 2020.

\bibitem{huang2019ccnet}
Zilong Huang, Xinggang Wang, Lichao Huang, Chang Huang, Yunchao Wei, and Wenyu
  Liu.
\newblock Ccnet: Criss-cross attention for semantic segmentation.
\newblock In {\em Proceedings of the IEEE/CVF international conference on
  computer vision}, pages 603--612, 2019.

\bibitem{larsson2016learning}
Gustav Larsson, Michael Maire, and Gregory Shakhnarovich.
\newblock Learning representations for automatic colorization.
\newblock In {\em Computer Vision--ECCV 2016: 14th European Conference,
  Amsterdam, The Netherlands, October 11--14, 2016, Proceedings, Part IV 14},
  pages 577--593. Springer, 2016.

\bibitem{li2023blip}
Junnan Li, Dongxu Li, Silvio Savarese, and Steven Hoi.
\newblock Blip-2: Bootstrapping language-image pre-training with frozen image
  encoders and large language models.
\newblock {\em arXiv preprint arXiv:2301.12597}, 2023.

\bibitem{li2020consistent}
Peike Li, Yunchao Wei, and Yi Yang.
\newblock Consistent structural relation learning for zero-shot segmentation.
\newblock {\em Advances in Neural Information Processing Systems},
  33:10317--10327, 2020.

\bibitem{li2021pseudo}
Yi Li, Zhanghui Kuang, Liyang Liu, Yimin Chen, and Wayne Zhang.
\newblock Pseudo-mask matters in weakly-supervised semantic segmentation.
\newblock In {\em Proceedings of the IEEE/CVF International Conference on
  Computer Vision}, pages 6964--6973, 2021.

\bibitem{coco}
Tsung-Yi Lin, Michael Maire, Serge Belongie, James Hays, Pietro Perona, Deva
  Ramanan, Piotr Doll{\'a}r, and C~Lawrence Zitnick.
\newblock Microsoft coco: Common objects in context.
\newblock In {\em Computer Vision--ECCV 2014: 13th European Conference, Zurich,
  Switzerland, September 6-12, 2014, Proceedings, Part V 13}, pages 740--755.
  Springer, 2014.

\bibitem{liu2020part}
Yongfei Liu, Xiangyi Zhang, Songyang Zhang, and Xuming He.
\newblock Part-aware prototype network for few-shot semantic segmentation.
\newblock In {\em Computer Vision--ECCV 2020: 16th European Conference,
  Glasgow, UK, August 23--28, 2020, Proceedings, Part IX 16}, pages 142--158.
  Springer, 2020.

\bibitem{long2015fully}
Jonathan Long, Evan Shelhamer, and Trevor Darrell.
\newblock Fully convolutional networks for semantic segmentation.
\newblock In {\em Proceedings of the IEEE conference on computer vision and
  pattern recognition}, pages 3431--3440, 2015.

\bibitem{lu2021simpler}
Zhihe Lu, Sen He, Xiatian Zhu, Li Zhang, Yi-Zhe Song, and Tao Xiang.
\newblock Simpler is better: Few-shot semantic segmentation with classifier
  weight transformer.
\newblock In {\em Proceedings of the IEEE/CVF International Conference on
  Computer Vision}, pages 8741--8750, 2021.

\bibitem{segclip}
Huaishao Luo, Junwei Bao, Youzheng Wu, Xiaodong He, and Tianrui Li.
\newblock Segclip: Patch aggregation with learnable centers for open-vocabulary
  semantic segmentation.
\newblock {\em arXiv preprint arXiv:2211.14813}, 2022.

\bibitem{context}
Roozbeh Mottaghi, Xianjie Chen, Xiaobai Liu, Nam-Gyu Cho, Seong-Whan Lee, Sanja
  Fidler, Raquel Urtasun, and Alan Yuille.
\newblock The role of context for object detection and semantic segmentation in
  the wild.
\newblock In {\em Proceedings of the IEEE conference on computer vision and
  pattern recognition}, pages 891--898, 2014.

\bibitem{nguyen2019feature}
Khoi Nguyen and Sinisa Todorovic.
\newblock Feature weighting and boosting for few-shot segmentation.
\newblock In {\em Proceedings of the IEEE/CVF International Conference on
  Computer Vision}, pages 622--631, 2019.

\bibitem{noroozi2016unsupervised}
Mehdi Noroozi and Paolo Favaro.
\newblock Unsupervised learning of visual representations by solving jigsaw
  puzzles.
\newblock In {\em Computer Vision--ECCV 2016: 14th European Conference,
  Amsterdam, The Netherlands, October 11-14, 2016, Proceedings, Part VI}, pages
  69--84. Springer, 2016.

\bibitem{pathak2016context}
Deepak Pathak, Philipp Krahenbuhl, Jeff Donahue, Trevor Darrell, and Alexei~A
  Efros.
\newblock Context encoders: Feature learning by inpainting.
\newblock In {\em Proceedings of the IEEE conference on computer vision and
  pattern recognition}, pages 2536--2544, 2016.

\bibitem{pinheiro2015image}
Pedro~O Pinheiro and Ronan Collobert.
\newblock From image-level to pixel-level labeling with convolutional networks.
\newblock In {\em Proceedings of the IEEE conference on computer vision and
  pattern recognition}, pages 1713--1721, 2015.

\bibitem{clip}
Alec Radford, Jong~Wook Kim, Chris Hallacy, Aditya Ramesh, Gabriel Goh,
  Sandhini Agarwal, Girish Sastry, Amanda Askell, Pamela Mishkin, Jack Clark,
  et~al.
\newblock Learning transferable visual models from natural language
  supervision.
\newblock In {\em International Conference on Machine Learning}, pages
  8748--8763. PMLR, 2021.

\bibitem{strudel2021segmenter}
Robin Strudel, Ricardo Garcia, Ivan Laptev, and Cordelia Schmid.
\newblock Segmenter: Transformer for semantic segmentation.
\newblock In {\em Proceedings of the IEEE/CVF international conference on
  computer vision}, pages 7262--7272, 2021.

\bibitem{yfcc}
Bart Thomee, David~A Shamma, Gerald Friedland, Benjamin Elizalde, Karl Ni,
  Douglas Poland, Damian Borth, and Li-Jia Li.
\newblock Yfcc100m: The new data in multimedia research.
\newblock {\em Communications of the ACM}, 59(2):64--73, 2016.

\bibitem{tian2020prior}
Zhuotao Tian, Hengshuang Zhao, Michelle Shu, Zhicheng Yang, Ruiyu Li, and Jiaya
  Jia.
\newblock Prior guided feature enrichment network for few-shot segmentation.
\newblock {\em IEEE transactions on pattern analysis and machine intelligence},
  44(2):1050--1065, 2020.

\bibitem{touvron2021training}
Hugo Touvron, Matthieu Cord, Matthijs Douze, Francisco Massa, Alexandre
  Sablayrolles, and Herv{\'e} J{\'e}gou.
\newblock Training data-efficient image transformers \& distillation through
  attention.
\newblock In {\em International conference on machine learning}, pages
  10347--10357. PMLR, 2021.

\bibitem{vaswani2017attention}
Ashish Vaswani, Noam Shazeer, Niki Parmar, Jakob Uszkoreit, Llion Jones,
  Aidan~N Gomez, {\L}ukasz Kaiser, and Illia Polosukhin.
\newblock Attention is all you need.
\newblock {\em Advances in neural information processing systems}, 30, 2017.

\bibitem{wang2018non}
Xiaolong Wang, Ross Girshick, Abhinav Gupta, and Kaiming He.
\newblock Non-local neural networks.
\newblock In {\em Proceedings of the IEEE conference on computer vision and
  pattern recognition}, pages 7794--7803, 2018.

\bibitem{Wang_2020_CVPR}
Yude Wang, Jie Zhang, Meina Kan, Shiguang Shan, and Xilin Chen.
\newblock Self-supervised equivariant attention mechanism for weakly supervised
  semantic segmentation.
\newblock In {\em Proceedings of the IEEE/CVF Conference on Computer Vision and
  Pattern Recognition (CVPR)}, June 2020.

\bibitem{xian2019semantic}
Yongqin Xian, Subhabrata Choudhury, Yang He, Bernt Schiele, and Zeynep Akata.
\newblock Semantic projection network for zero-and few-label semantic
  segmentation.
\newblock In {\em Proceedings of the IEEE/CVF Conference on Computer Vision and
  Pattern Recognition}, pages 8256--8265, 2019.

\bibitem{groupvit}
Jiarui Xu, Shalini De~Mello, Sifei Liu, Wonmin Byeon, Thomas Breuel, Jan Kautz,
  and Xiaolong Wang.
\newblock Groupvit: Semantic segmentation emerges from text supervision.
\newblock In {\em Proceedings of the IEEE/CVF Conference on Computer Vision and
  Pattern Recognition}, pages 18134--18144, 2022.

\bibitem{Xu_2021_ICCV}
Lian Xu, Wanli Ouyang, Mohammed Bennamoun, Farid Boussaid, Ferdous Sohel, and
  Dan Xu.
\newblock Leveraging auxiliary tasks with affinity learning for weakly
  supervised semantic segmentation.
\newblock In {\em Proceedings of the IEEE/CVF International Conference on
  Computer Vision (ICCV)}, pages 6984--6993, October 2021.

\bibitem{xu2022simple}
Mengde Xu, Zheng Zhang, Fangyun Wei, Yutong Lin, Yue Cao, Han Hu, and Xiang
  Bai.
\newblock A simple baseline for open-vocabulary semantic segmentation with
  pre-trained vision-language model.
\newblock In {\em Computer Vision--ECCV 2022: 17th European Conference, Tel
  Aviv, Israel, October 23--27, 2022, Proceedings, Part XXIX}, pages 736--753.
  Springer, 2022.

\bibitem{yang2021mining}
Lihe Yang, Wei Zhuo, Lei Qi, Yinghuan Shi, and Yang Gao.
\newblock Mining latent classes for few-shot segmentation.
\newblock In {\em Proceedings of the IEEE/CVF international conference on
  computer vision}, pages 8721--8730, 2021.

\bibitem{yuan2021florence}
Lu Yuan, Dongdong Chen, Yi-Ling Chen, Noel Codella, Xiyang Dai, Jianfeng Gao,
  Houdong Hu, Xuedong Huang, Boxin Li, Chunyuan Li, et~al.
\newblock Florence: A new foundation model for computer vision.
\newblock {\em arXiv preprint arXiv:2111.11432}, 2021.

\bibitem{yuan2021ocnet}
Yuhui Yuan, Lang Huang, Jianyuan Guo, Chao Zhang, Xilin Chen, and Jingdong
  Wang.
\newblock Ocnet: Object context for semantic segmentation.
\newblock {\em International Journal of Computer Vision}, 129(8):2375--2398,
  2021.

\bibitem{zhao2017open}
Hang Zhao, Xavier Puig, Bolei Zhou, Sanja Fidler, and Antonio Torralba.
\newblock Open vocabulary scene parsing.
\newblock In {\em Proceedings of the IEEE International Conference on Computer
  Vision}, pages 2002--2010, 2017.

\bibitem{zhao2017pyramid}
Hengshuang Zhao, Jianping Shi, Xiaojuan Qi, Xiaogang Wang, and Jiaya Jia.
\newblock Pyramid scene parsing network.
\newblock In {\em Proceedings of the IEEE conference on computer vision and
  pattern recognition}, pages 2881--2890, 2017.

\bibitem{zheng2021rethinking}
Sixiao Zheng, Jiachen Lu, Hengshuang Zhao, Xiatian Zhu, Zekun Luo, Yabiao Wang,
  Yanwei Fu, Jianfeng Feng, Tao Xiang, Philip~HS Torr, et~al.
\newblock Rethinking semantic segmentation from a sequence-to-sequence
  perspective with transformers.
\newblock In {\em Proceedings of the IEEE/CVF conference on computer vision and
  pattern recognition}, pages 6881--6890, 2021.

\bibitem{zhou2021denseclip}
Chong Zhou, Chen~Change Loy, and Bo Dai.
\newblock Denseclip: Extract free dense labels from clip.
\newblock {\em arXiv preprint arXiv:2112.01071}, 2021.

\bibitem{zhu2023minigpt}
Deyao Zhu, Jun Chen, Xiaoqian Shen, Xiang Li, and Mohamed Elhoseiny.
\newblock Minigpt-4: Enhancing vision-language understanding with advanced
  large language models.
\newblock {\em arXiv preprint arXiv:2304.10592}, 2023.

\end{thebibliography}
